\pdfoutput=1
\documentclass{article}

\usepackage{microtype}
\usepackage{graphicx}
\usepackage{subfigure}
\usepackage{booktabs} 
\usepackage{mathtools}
\usepackage{amssymb}
\usepackage{amsmath}
\usepackage{bm}
\usepackage{nicefrac}       
\usepackage{hyperref}       

\usepackage{amsthm}
\newtheorem{prop}{Proposition}

\newtheorem{thm}{Theorem}

\usepackage{color}
\usepackage{xcolor}
\usepackage{lipsum}

\usepackage{hyperref}

\usepackage[accepted]{icml2020}

\icmltitlerunning{Predict then Interpolate: A Simple Algorithm to Learn Stable Classifiers}

\begin{document}

\twocolumn[
\icmltitle{Predict then Interpolate: A Simple Algorithm to Learn Stable Classifiers}


\icmlsetsymbol{equal}{*}

\begin{icmlauthorlist}
\icmlauthor{Yujia Bao}{yb}
\icmlauthor{Shiyu Chang}{sc}
\icmlauthor{Regina Barzilay}{rb}
\end{icmlauthorlist}

\icmlaffiliation{yb}{MIT CSAIL}
\icmlaffiliation{sc}{MIT-IBM Watson AI Lab}
\icmlaffiliation{rb}{MIT CSAIL}

\icmlcorrespondingauthor{Yujia Bao}{yujia@csail.mit.edu}

\icmlkeywords{Machine Learning, ICML}

\vskip 0.3in
]



\printAffiliationsAndNotice{}  

\begin{abstract}
We propose Predict then Interpolate (\textsc{pi}), a simple algorithm for learning correlations that are stable across environments. The algorithm follows from the intuition that when using a classifier trained on one environment to make predictions on examples from another environment, its mistakes are informative as to which correlations are unstable. In this work, we prove that by interpolating the distributions of the correct predictions and the wrong predictions, we can uncover an oracle distribution where the unstable correlation vanishes. Since the oracle interpolation coefficients are not accessible, we use group distributionally robust optimization to minimize the worst-case risk across all such interpolations. We evaluate our method on both text classification and image classification. Empirical results demonstrate that our algorithm is able to learn robust classifiers (outperforms IRM by 23.85\% on synthetic environments and 12.41\% on natural environments). Our code and data are available at \url{https://github.com/YujiaBao/Predict-then-Interpolate}.
\end{abstract}

\label{submission}

\section{Introduction}\label{sec:intro}

Distributionally robust optimization (DRO) alleviates model biases by minimizing the worst-case risk over a set of human-defined groups. However, in order to construct these groups, humans must identify and annotate these biases, a process as expensive as annotating the label itself~\cite{ben2013robust,duchi2018learning,sagawa2019distributionally}. In this paper we propose a simple algorithm to create groups that are informative of these biases, and use these groups to train stable classifiers.

Our algorithm operates on data split among multiple environments, across which correlations between bias features and the label may vary. Instead of handcrafting environments based on explicit, task-dependent biases, these environments can be determined by generic information that is easy to collect~\cite{peters2015causal}. For example, environments can represent data collection circumstances, like location and time. Our goal is to learn correlations that are stable across these environments.

Given these environments, one could directly use them as groups for DRO. Doing so would optimize the worst-case risk over all interpolations of the training environments. However, if the unstable (bias) features are positively \emph{and} differentially correlated with the label in all training environments, the unstable correlation will be positive in any of their interpolations. DRO, optimizing for the best worst-case performance, will inevitably exploit these unstable features, and we fail to learn a stable classifier.

In this work, we propose Predict then Interpolate (\textsc{pi}), a simple recipe for creating groups whose interpolation yields a distribution with only stable correlations.
Our idea follows from the intuition that when using a classifier trained on one environment to make predictions on examples from a different environment,
its mistakes are informative of the unstable correlations.
In fact, we can prove that if the unstable features and the label are positively correlated across all environments, the same correlation flips to negative in the set of mistakes. Therefore, by interpolating the distributions of correct and incorrect predictions, we can uncover an ``oracle'' distribution in which only stable features are correlated with the label. Although the oracle interpolation coefficients are not accessible, we can minimize the worst-case risk over all interpolations, providing an upper bound of the risk on the oracle distribution.

Our learning paradigm consists of three steps. First, we train an individual classifier for each environment to estimate the conditional distribution of the label given the input. These classifiers are biased, as they may rely on any correlations in the dataset. Next, we apply each environment's classifier to partition all other environments, based on prediction correctness. Finally, we obtain our robust classifier by minimizing the worst-case risk over all interpolations of the partitions.

Empirically, we evaluate our approach on both synthetic and real-world environments.
First, we simulate unstable correlations in synthetic environments by appending spurious features. Our results in both digit classification and aspect-level sentiment classification demonstrate that our method delivers significant performance gain (23.85\% absolute accuracy) over invariant risk minimization (IRM), approaching oracle performance. Quantitative analyses confirm that our method generates partitions with opposite unstable correlations. Next, we applied our approach on natural environments defined by an existing attribute of the input. Our experiments on CelebA and ASK2ME showed that directly applying DRO on environments improves robust accuracy for known attributes, but this robustness doesn't generalize equally across other attributes that are unknown during train time. On the other hand, by creating partitions with opposite unstable correlations, our method is able to improve average worst-group accuracy by 12.41\% compared to IRM.

\section{Related work}\label{sec:related}

\paragraph{Removing known biases:} Large scale datasets are fraught with biases.
For instance, in face recognition~\cite{liu2015deep},
spurious associations may exist between different face attributes (e.g. hair color)
and demographic information (e.g. ethnicity)~\cite{buolamwini2018gender}.
Furthermore, in natural language inference~\cite{bowman-etal-2015-large},
the entailment label can often be predicted from lexical overlap of the two inputs~\cite{mccoy-etal-2019-right}.
Finally, in molecular property prediction~\cite{wu2018moleculenet,mayr2018large}, performance varies significantly across different scaffolds~\cite{yang2019analyzing}.

Many approaches have been proposed to mitigate biases
when they are known beforehand.
Examples include adversarial training to remove biases from representations~\cite{belinkov2019don,stacey-etal-2020-avoiding},
re-weighting training examples~\cite{schuster2019towards}, and
combining a biased model and the base model's predictions using a product of experts~\cite{hinton2002training,clark2019don,he2019unlearn, mahabadi2020end}.
These models are typically designed for a specific type of bias and thus require extra domain knowledge to generalize to new tasks.

Group DRO is another attractive framework since it allows explicit modeling of the
distribution family that we want to optimize over.
Previous work~\cite{hu2018does, oren2019distributionally, Sagawa*2020Distributionally}
has shown the effectiveness of group DRO to train un-biased models.
In these models, the groups are specified by human based on the knowledge of the
bias attributes.
Our work differs from them as we create groups using trained models.
This allows us to apply group DRO when we don't have annotations for the bias
attributes.
Moreover, when the bias attributes are available,
we can further refine our groups to reduce unknown biases.
{}
\paragraph{Removing unknown biases:} Determining dataset biases is time-consuming and often requires task-specific expert
knowledge~\cite{zellers-etal-2019-hellaswag,sakaguchi2020winogrande}.
Thus, there are two lines of work that aim to build robust models without explicitly knowing the type of bias.
The first assumes that weak models, which have limited capacity~\cite{sanh2021learning} or are
under-trained~\cite{utama2020towards}, are more prone to rely on shallow heuristics and
rediscover previously human-identified dataset biases.
By learning from the weak models' mistakes,
we can obtain a more robust model.
While these methods show empirical benefits on some NLP tasks, the extent to which their assumption holds is unclear.
In fact, recent work~\cite{sagawa2020overparameterization} shows that
over-parametrization may actually exacerbate unstable correlations for image classification.

The second line of work assumes that the training data are collected from
separate environments, across which unstable features exhibit different correlations with the label~\cite{peters2016causal,krueger2020outofdistribution,chang2020invariant,jin2020enforcing,ahuja2020invariant,arjovsky2019invariant}.
Invariant risk minimization~\cite{arjovsky2019invariant},
a representative method along this line, learns representations that are simultaneously optimal across all environments.
However, since this representation is trained across all environments,
it can easily degenerate in real-world applications~\cite{gulrajani2020search}.
One can consider an extreme case where the learned representation directly
encodes the one-hot embedding of the label.
While this learned representation is stable (invariant) according to the definition,
the model can utilize \emph{any unstable features} to generate this
representation.
We have no guarantee on how the model would generalize when the unstable
correlations vanish.

Our algorithm instead decomposes the problem of learning stable classifiers into two parts: finding \emph{unstable features} and training a robust model. By constraining the classifiers to be environment-specific in the first part, we are able to construct an oracle distribution where the unstable features are not correlated with the label.
Our model then directly optimizes an upper bound of the risk on this oracle distribution.
Empirically, we demonstrate that our method is able to eliminate biases not given during training on multiple real-world applications.

\section{Method}\label{sec:method}
We consider the standard setting~\cite{arjovsky2019invariant}
where the training data are comprised of $n$ environments
$\mathcal{E} = \{E_1, \ldots, E_n\}$.
For each environment $E_i$, we have input-label pairs $(x, y) \overset{\text{iid}}{\sim} P_i$.
Our goal is to learn correlations that are \emph{stable}
across these environments~\cite{woodward2005making} so that the model can generalize to a new test environment $E_{\text{test}}$ that has the same stable correlations.

\subsection{Algorithm}
Our intuition follows from a simple question.

\emph{What happens if we apply a classifier $f_i$ trained on environment
$E_i$ to a different environment $E_j$?}

Suppose we have enough data in $E_i$ and the classifier $f_i$ is able to
perfectly fit the underlying conditional $P_i(y | x)$.
Since $E_i$ and $E_j$ follow different distributions,
the classifier $f_i$ will make mistakes on $E_j$.
These mistakes are natural products of the unstable correlation:
if the correlation of the unstable feature is higher in $E_i$ than in $E_j$,
the classifier $f_i$ will overuse this feature when making predictions in $E_j$.

In fact, we can show that under certain conditions,
the unstable correlation within the subset of wrong
predictions is opposite of that within the subset of
correct predictions (Section~\ref{sec:theory}).
By interpolating between these two subsets,
we can uncover an \emph{oracle distribution} where the label is not correlated with
the unstable feature.
Since this interpolation coefficient is not accessible in practice,
we adopt group DRO to minimize the worst-case risk over
all interpolations of these subsets.
This provides us an upper bound of the risk on the
oracle distribution.

Concretely, our approach has three successive stages.
\begin{description}
  \item[Stage 1:] For each environment $E_i$, train an environment specific classifier
    $f_i$.
  \item[Stage 2:] For each pair of environments $E_i$ and $E_j$,
    use the trained classifier $f_i$ to partition $E_j$ into two sets
    \[
      E_j = E_{j}^{i \checkmark} \cup E_j^{i \times}
    \]
    where $E_{j}^{i \checkmark}$ contains examples that $f_i$ predicted correctly and $E_j^{i \times}$ contains those predicted incorrectly.
  \item[Stage 3:] Train the final model $f$ by minimizing the worst-case risk over
    the set of all interpolations $\mathcal{Q}$:
    \[
      \mathcal{Q} = \left\{
        \sum_{i \neq j} \lambda_j^{i \checkmark} P_j^{i \checkmark}
        + \lambda_j^{i \times} P_j^{i \times} :
        \sum_{i \neq j} \lambda_j^{i \checkmark} + \lambda_j^{i \times} = 1
      \right\},
    \]
    where $P_j^{i \checkmark}$ and $P_j^{i \times}$ are the
    empirical distributions of $E_{j}^{i \checkmark}$ and $E_j^{i \times}$.
    Because the optimum value of a linear program must occur at a vertex,
    the worst-case risk over $\mathcal{Q}$ is equivalent to the maximum expected
    risk across all groups.
    This allows us to formulate the objective as a min-max problem:
    \[
      \min_{f} \max_{P \in \mathcal{P}}
      \mathbb{E}_{(x, y) \sim P} [\mathcal{L}(x, y; f)],
    \]
    where $\mathcal{L}(x, y; f)$ is the loss of the model $f$ and
    $\mathcal{P}$ is the set of distributions
    $\{P_j^{i \checkmark}\}_{i\neq j} \cup \{P_j^{i \times}\}_{i \neq j}$.
\end{description}
\paragraph{Extensions of the algorithm:}
For regression tasks,
we can set a threshold on the mean square error to partition environments.
We can also apply the first two stages multiple times, treating new partitions as different environments, to iteratively refine the groups.
In this work, we focus on the basic setting and leave the rest for future work.

\begin{figure*}[t]
  \centering
  \includegraphics[width=\linewidth]{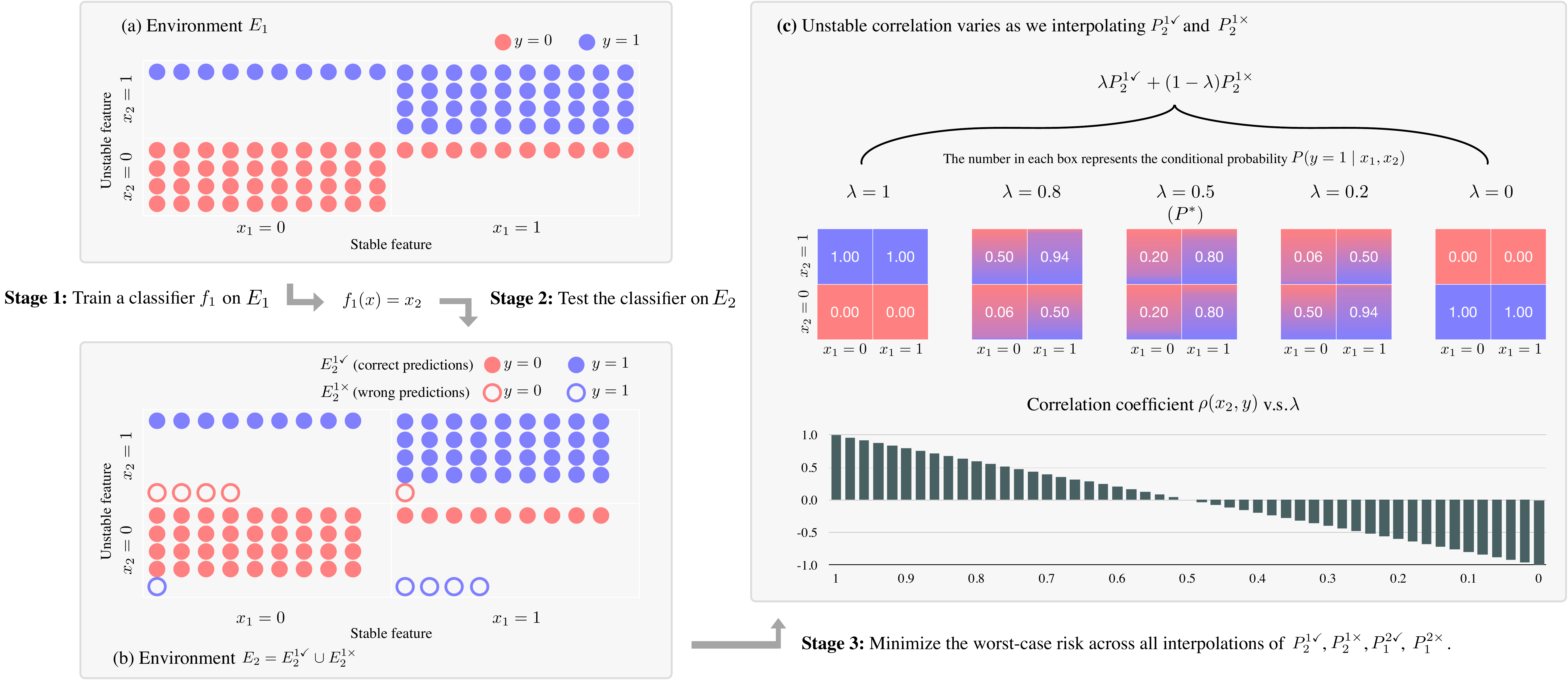}
  \vspace{-5mm}
  \caption{Illustration of our algorithm on the toy example.
    The label $y$ agrees with the stable feature $x_1$ with probability $0.8$ on both environments.
    For the unstable feature $x_2$, the probability of $x_2=y$ is $1.0$ in $E_1$ and $0.9$ in $E_2$.
    Stage 1: We train a classifier $f_1$ on $E_1$.
    It learns to make predictions solely based on the unstable feature $x_2$.
    Stage 2:
    We use $f_1$ to partition $E_2$ based on the prediction correctness.
    While the correlation of $x_2$ is positive for both $E_1$ and $E_2$,
    it flips to negative in set of wrong predictions $E_2^{1\times}$.
    Stage 3:
    Interpolating $P_2^{1\checkmark}$ and $P_2^{1\times}$ allows us to uncover
    an oracle distribution $P^*$ where the unstable feature $x_2$ is not
    correlated with the label.
    Note that here we only illustrate how to partition $E_2$ using $f_1$.
    In our algorithm, we also use the classifier $f_2$ (trained on $E_2$) to partition $E_1$,
    and the final model $f$ is obtained by minimizing the worst-case risk over all
    interpolations of $P_1^{2\checkmark}, P_1^{2\times}, P_2^{1\checkmark}P_2^{1\times}$.
  }\label{fig:toy}
  \vspace{-4mm}
\end{figure*}

\subsection{A toy example}
\label{sec:toy}
To understand the behavior of the algorithm,
let's consider a simple example with two environments $E_1$ and $E_2$ (Figure~\ref{fig:toy}).
For each environment, the data are generated by the following
process.\footnote{\citet{arjovsky2019invariant} used this process to
construct the Colored-MNIST dataset.}
\begin{itemize}
  \item First, sample the feature $x_1 \in \{0, 1\}$ which
    takes the value $1$ with probability $0.5$. This is our stable feature.
  \item Next, sample the observed noisy label $y \in \{0, 1\}$ by flipping the value of $x_1$
    with probability $0.2$.
  \item Finally, for each environment $E_i$, sample the unstable feature $x_2
    \in \{0, 1\}$
    by flipping the value of $y$ with probability $\eta_i$.
    Let $\eta_1 = 0$ and $\eta_2 = 0.1$.
\end{itemize}
Our goal is to learn a classifier that only uses feature $x_1$ to predict $y$.
Since the unstable feature $x_2$ is positively correlated with the label
across both environments,
directly treating the environments as groups and applying group DRO
will also exploit this correlation during training.

Let's take a step back and consider a classifier $f_1$ that is trained only on
$E_1$.
Since $x_2$ is identical to $y$ and $x_1$ differs from $y$ with probability
$0.2$, $f_1$ simply learns to ignore $x_1$ and predict $y$ as $x_2$
(Figure~\ref{fig:toy}a).
When we apply $f_1$ to the other environment $E_2$,
it will make mistakes on examples where $x_2$ is flipped from $y$.
Moreover, we can check that the correlation coefficient between the unstable
feature $x_2$ and $y$ is $1$ in the set of correct predictions
$E_2^{1 \checkmark}$ and it flips to $-1$ in the set of mistakes $E_2^{1
\times}$ (Figure~\ref{fig:toy}b).
In this toy example, the \emph{oracle distribution} $P^*$,
where the correlation between $x_2$ and $y$ is $0$,
can be obtained by simply averaging the empirical distribution of the two
subsets (Figure~\ref{fig:toy}c):
\[
  P^*(x_1, x_2, y) = 0.5 P_2^{1 \checkmark}(x_1, x_2, y) + 0.5
  P_2^{1\times}(x_1, x_2, y).
\]
We can also verify that the optimal solution that minimize the worst-case risk
across $E_{2}^{1 \checkmark}$ and $E_2^{1 \times}$ is to predict $y$ only using
$x_1$. (Appendix~\ref{app:toy_verify}).

\paragraph{Remark 1:}
In the algorithm, we also use the classifier $f_2$ trained on $E_2$ to partition $E_1$. The final model $f$ is obtained by minimizing
the worst-case risk over $P_1^{2\checkmark}, P_1^{2\times}, P_2^{1\checkmark}, P_2^{1\times}$.
\paragraph{Remark 2:}
Our algorithm optimizes an \emph{upper bound} of the risk on the oracle distribution.
In general, it \emph{doesn't guarantee} that the unstable correlation is not utilized
by the model when the worst-case performance is not achieved at the oracle distribution.

\subsection{Theoretical analysis}
\label{sec:theory}
In the previous example,
we have seen that the unstable correlation flips in the set of mistakes
$E_2^{1\times}$ compared to the set of correct predictions $E_2^{1\checkmark}$.
Here, we would like to investigate how
this property holds in general.\footnote{All proofs are relegated to
Appendix~\ref{app:theory}.}
We focus our analysis on binary classification tasks where $y \in \{0, 1\}$.
Let $x_1$ be the stable feature and $x_2$ be unstable feature that has
various correlations across environments.
We use capital letters $X_1, X_2, Y$ to represent random variables
and use lowercase letters $x_1, x_2, y$ to denote their specific values.

\begin{prop}
  \label{prop:1}
  For a pair of environments $E_i$ and $E_j$,
  assuming that the classifier $f_i$ is able to learn the true conditional
  $P_i(Y\mid X_1, X_2)$,
  we can write the joint distribution $P_j$ of $E_j$ as the mixture of
  $P_j^{i \checkmark}$ and $P_j^{i \times}$:
  \[
    P_j(x_1, x_2, y) = \alpha_j^i P_j^{i \checkmark} (x_1, x_2, y) +
    (1-\alpha_j^i)  P_j^{i \times} (x_1, x_2, y),
  \]
  where $\alpha_j^i = \sum_{x_1, x_2, y} P_j(x_1, x_2, y)\cdot P_i(y \mid x_1, x_2)$ and
  \[
    \begin{aligned}
      P_j^{i\checkmark}(x_1, x_2, y) &\propto
      P_j(x_1, x_2, y)\cdot P_i(y \mid x_1, x_2),\\
      P_j^{i\times}(x_1, x_2, y) &\propto
      P_j(x_1, x_2, y)\cdot P_i(1-y \mid x_1, x_2).
    \end{aligned}
  \]
\end{prop}

Intuitively, when partitioning the environment $E_j$,
we are scaling its joint distribution based on the conditional on $E_i$.

\paragraph{Two degenerate cases:} From Proposition~\ref{prop:1}, we see that
the algorithm degenerates when $\alpha_j^i = 0$ (predictions of $f_i$ are all
wrong) or $\alpha_j^i = 1$ (predictions of $f_i$ are all correct).
The first case occurs when the unstable correlation is flipped between $P_i$ and
$P_j$.
One may think about setting $\eta_1 = 0$ and $\eta_2 = 1$ in the toy example.
In this case, we can obtain the oracle distribution by directly interpolating $P_i$
and $P_j$.
The second case implies that the conditional is the same across the environments:
$P_i(Y \mid X_1, X_2) = P_j(Y \mid X_1, X_2)$.
Since $x_2$ is the unstable feature, this equality holds when
the conditional mutual information between $X_2$ and $Y$ is zero given $X_1$,
i.e., $P_i(Y \mid X_1, X_2) = P_i(Y \mid X_1)$.
In this case, $f_i$ already ignores the unstable feature $x_2$.

To carryout the following analysis, we assume that the marginal distribution of
$Y$ is uniform in all joint distributions,
i.e., $f_i$ performs equally well on different labels.

\begin{thm}
  \label{thm:1}
  Suppose $X_2$ is independent of $X_1$ given $Y$.
  For any environment pair $E_i$ and $E_j$,
  if $\sum_y P_i(x_2 \mid y) = \sum_y P_j(x_2 \mid y)$ for any $x_2$,
  then $\mathrm{Cov}(X_2, Y; P_i) > \mathrm{Cov}(X_2, Y; P_j)$ implies
  $\mathrm{Cov}(X_2, Y; P_j^{i\times}) < 0$ and
  $\mathrm{Cov}(X_2, Y; P_i^{j\times}) > 0$.
\end{thm}
The result follows from the connection between the covariance and the conditional.
On one side, the covariance between $x_2$ and $Y$ captures the difference of
their conditionals: $P(X_2 \mid Y = 1) - P(X_2 \mid Y = 0)$,
On the other side, the conditional independence assumption allows us to factorize
the joint distribution:
$P_i(x_1, x_2, y) = P_i(x_1, y) P_i(x_2 \mid y)$.
Combining them together finishes the proof.

Theorem~\ref{thm:1} tells us no matter whether the spurious correlation is positive or negative,
we can obtain an oracle distribution $P^*$, 
$\mathrm{Cov}(X_2, Y; P^*) = 0$ by interpolating across
$P_j^{i \checkmark}$, $P_j^{i \times}$, $P_i^{j \checkmark}$, $P_i^{j \times}$.
By optimizing the worst-case risk across all interpolations,
our final model $f$ provides an \emph{upper bound} of the risk on the oracle
distribution $P^*$.

We also note that the toy example in Section~\ref{sec:toy} is a special case
of the assumption in Theorem~\ref{thm:1}.
While many previous work also construct datasets with this
assumption~\cite{arjovsky2019invariant, choe2020empirical},
it may be too restrictive in practice.
In the general case,
although we cannot guarantee the sign of the correlation,
we can still obtain an upper bound for $\mathrm{Cov}(X_2, Y; P_j^{i\times})$
and a lower bound for $\mathrm{Cov}(X_2, Y; P_i^{j\times})$:

\begin{thm}
  \label{thm:2}
  For any environment pair $E_i$ and $E_j$,
  $\mathrm{Cov}(X_2, Y; P_i) > \mathrm{Cov}(X_2, Y; P_j)$ implies
  \[
    \begin{aligned}
& \mathrm{Cov}(X_2, Y; P_j^{i\times})\\
&< \frac{1-\alpha_j^i}{\alpha_{i}^{i}} \mathrm{Cov}(X_2, Y; P_i^{i\checkmark}) - 
\frac{1-\alpha_j^i}{\alpha_j^i} \mathrm{Cov}(X_2, Y; P_j^{i\checkmark})\\
& \mathrm{Cov}(X_2, Y; P_i^{j\times})\\
&> \frac{1-\alpha_i^j}{\alpha_j^j} \mathrm{Cov}(X_2, Y; P_j^{j\checkmark}) - 
\frac{1-\alpha_i^j}{\alpha_i^j}\mathrm{Cov}(X_2, Y; P_i^{j\checkmark})
    \end{aligned}
  \]
  where $P_i^{i\checkmark}$ is the distribution of the correct predictions when
  applying $f_i$ on $E_i$.
\end{thm}

Intuitively, if the correlation is stronger in $E_i$,
then the classifier $f_i$ will overuse this correlation and make mistakes on $E_j$
when this stronger correlation doesn't hold.
Conversely, the classifier $f_j$ will underuse this correlation and make
mistakes on $E_i$ when the correlation is stronger.

\begin{table*}[h]
  \centering
  \begin{tabular}{lcccccccccccc}
  \toprule
  &
  \multicolumn{2}{c}{\textsc{erm}} &
  \multicolumn{2}{c}{\textsc{dro}} &
  \multicolumn{2}{c}{\textsc{irm}} &
  \multicolumn{2}{c}{\textsc{rgm}} &
  \multicolumn{2}{c}{\textsc{pi (ours)}} &
  \multicolumn{2}{c}{\textsc{oracle}}
  \\
  \cmidrule(lr{0.5em}){2-3}
  \cmidrule(lr{0.5em}){4-5}
  \cmidrule(lr{0.5em}){6-7}
  \cmidrule(lr{0.5em}){8-9}
  \cmidrule(lr{0.5em}){10-11}
  \cmidrule(lr{0.5em}){12-13}
  $P_{\text{val}} = P_{\text{test}}$?
  & \checkmark & $\times$
  & \checkmark & $\times$
  & \checkmark & $\times$
  & \checkmark & $\times$
  & \checkmark & $\times$
  & \checkmark & $\times$
  \\
  \midrule
  MNIST    & 26.15 & 14.25 & 32.51 & 21.06 & 45.41 &13.13 &   42.49 & 15.33 &
              $\bm{69.44}$ & $\bm{69.68}$ & 71.40 & 71.60 \\
  \midrule
			Beer Look   & 64.63 & 60.96 & 64.53 & 62.75 & 65.83 & 63.31  & 66.31  & 61.51&
              $\bm{78.09}$ & $\bm{70.66}$ & 80.32 & 73.51 \\

  \midrule
  Beer Aroma & 55.25 & 51.99 & 57.08 & 53.39 & 60.25  & 53.25 &  66.33 &57.91
             & $\bm{77.01}$ & $\bm{67.35}$ & 77.34 & 69.99 \\
  \midrule
  Beer Palate & 49.01 & 46.69 & 47.72 & 46.35 & 66.45  & 44.09 & 68.77  & 44.81
              & $\bm{74.14}$ & $\bm{61.51}$ & 74.89 & 66.33 \\
  \bottomrule
  \end{tabular}
  \caption{
    Accuracy of different methods on image classification (majority baseline
    10\%) and aspect-level sentiment classification (majority baseline 50\%).
    All methods are tuned based on a held-out validation set.
    We consider two validation settings: 1) sample the validation set from the
    testing environment ($P_{\text{val}} = P_{\text{test}}$);
    2) sample the validation set from the training environment.
  }\label{tab:syn}
  \vspace{-3mm}
\end{table*}

\section{Experimental setup}\label{sec:setup}

\subsection{Datasets and Settings}
\paragraph{Synthetic environments:}
To assess the empirical behavior of our algorithm,
we start with controlled experiments where we can simulate spurious
correlation.
We consider two standard datasets: MNIST~\cite{lecun1998gradient} and
BeerReview~\cite{mcauley2012learning}.\footnote{All dataset statistics are
relegated to Appendix~\ref{app:dataset}.}

For MNIST, we adopt \citet{arjovsky2019invariant}'s approach for generating
spurious correlation and extend it to a more challenging multi-class problem.
For each image,
we sample $y$, which takes on the same value as its numeric digit with 0.75
probability and a uniformly random other digit with the remaining probability.
The spurious feature in sampled in a similar way:
it takes on the same value as $y$ with $\eta$ probability and a uniformly
random other value with the remaining probability.
We color the image according to the value of the spurious feature.
We set $\eta$ to $0.9$ and $0.8$ respectively for the training environments
$E_1$ and $E_2$.
In the testing environment, $\eta$ is set to $0.1$.

For BeerReview, we consider three aspect-level sentiment classification
tasks: look, aroma and palate~\cite{lei2016rationalizing,bao2018deriving}.
For each review, we append an artificial token (\texttt{art\_pos} or \texttt{art\_neg})
that is spuriously correlated with the binary sentiment label (\texttt{pos} or \texttt{neg}).
The artificial token agrees with the sentiment label with probability $0.9$ in
environment $E_1$ and with probability $0.8$ in environment $E_2$.
In the testing environment, the probability reduces to $0.1$.
Unlike MNIST, here we do not inject artificial label noise to the datasets.

Validation set plays a crucial role when the training distribution is different
from the testing distribution~\cite{gulrajani2020search}.
For both datasets, we consider two different validation settings and report
their performance separately:
1) sampling the validation set from the training environment;
2) sampling the validation set from the testing environment.

\paragraph{Natural environments:}
We also consider a practical setting where environments are naturally defined
by some attributes of the input and we want to use them to reduce biases that
are \emph{unknown} during training and validation.
We study two datasets: CelebA~\cite{liu2015faceattributes} where the
attributes are annotated by human and ASK2ME~\cite{doi:10.1200/CCI.19.00042} where the attributes are automatically generated by rules.

CelebA is an image classification dataset where each input image (face) is paired with 40 binary attributes.
We adopt \citet{sagawa2019distributionally}'s setting and treat hair color ($y
\in \{\texttt{blond}, \texttt{dark}\}$) as the target task.
We use the gender attribute to define the two training
environments, $E_1$=\{\texttt{female}\} and $E_2$=\{\texttt{male}\}.
Our goal is to learn a classifier that is robust to other unknown attributes such as \texttt{wearing\_hat}.
For model selection, we partition the validation data into four groups based on the
gender value and the label value: \{\texttt{female}, \texttt{blond}\}, \{\texttt{female}, \texttt{dark}\}, \{\texttt{male},
\texttt{blond}\}, \{\texttt{male}, \texttt{dark}\}.
We use the worst-group accuracy as our validation criteria.

ASK2ME is a text classification dataset where an input text (paper abstract from PubMed) is paired with 17 binary attributes, each indicating the presence of a different disease.
The task is to predict whether the input is informative about the \emph{risk} of cancer for gene mutation carriers, rather than cancer itself~\cite{deng2019validation}.
We define two training environments based on the \texttt{breast\_cancer} attribute,
$E_1$=\{\texttt{breast\_cancer=0}\} and $E_2$=\{\texttt{breast\_cancer=1}\}.
We would like to see whether the classifier is able to remove spurious correlations from other diseases that are unknown during training. Similar to CelebA, we compute the worst-group accuracy based on the \texttt{breast\_cancer} value and the label value and use it
for validation.

At test time, we evaluate the classifier's prediction robustness on all
attributes over a held-out test set.
For each attribute, we report the worst-group accuracy and the average-group
accuracy.

\subsection{Baselines}
  We compare our algorithm against the following baselines:

  \textbf{ERM}: We combine all environments together and apply standard empirical
  risk minimization.

  \textbf{IRM}: Invariant risk minimization~\cite{arjovsky2019invariant} learns a
  representation such that the linear classifier on top of this representation is
  simultaneously optimal across different environments.

  \textbf{RGM}: Regret minimization~\cite{jin2020enforcing} simulates unseen
  environments by using part of the training set as held-out environments.
  It quantifies the generalization ability in terms of regret, the difference
  between the losses of two auxiliary predictors trained with and without examples
  in the current environment.

  \textbf{DRO}: We can also apply DRO on groups defined by the environments and the labels.
  For example, in beer review,
  we can partition the training data into the four groups:
  \{\texttt{pos}, $E_1$\}, \{\texttt{neg}, $E_1$\}
  \{\texttt{pos}, $E_2$\}, \{\texttt{neg}, $E_2$\}.
  Minimizing the worst-case performance over these human-defined groups has shown success in improving model robustness
  \cite{sagawa2019distributionally}.
  
\textbf{Oracle}:
  In the synthetic environments,
  we can use the spurious features to define groups and train an oracle DRO model.
  For example, in beer review,
  the oracle model will minimize the worst-case risk over the four groups:
  \{\texttt{pos}, \texttt{art\_pos}\}, \{\texttt{pos}, \texttt{art\_neg}\}
  \{\texttt{pos}, \texttt{art\_pos}\}, \{\texttt{pos}, \texttt{art\_neg}\}.
  This helps us analyze the contribution of our algorithm in isolation of the
  inherent limitations of the task.

  For fair comparison, all methods share the same model architecture.\footnote{For IRM and RGM, in order to tune the weights and annealing strategy for the regularizer, the hyper-parameter search space is 21$\times$ larger than other methods.}
  Implementation details can be found in Appendix~\ref{app:details}.

\begin{table}[t]
  \centering
  \begin{tabular}{lcccc}
  \toprule
  & $E_1$ & $E_2$ & $E_2^{1\checkmark}$ & $E_2^{1\times}$ \\
  \midrule
    MNIST       & 0.8955 & 0.7769 & 0.9961 & $-0.1040$ \\
  \midrule
    Beer Look   & 0.8007 & 0.6006 & 0.8254 & $-0.8030$ \\
  \midrule
    Beer Aroma  & 0.8007 & 0.6006 & 0.9165 & $-0.9303$ \\
  \midrule
    Beer Palate & 0.8007 & 0.6006 & 0.9394 & $-0.9189$ \\
  \bottomrule
  \end{tabular}
  \caption{
    Pearson correlation coefficient between the spurious feature and the label
    across four datasets.
    While the correlation is positive for both training environments,
    it flips to negative in the set of wrong predictions $E_2^{1\times}$.
    Interpolating across $E_2^{1\checkmark}$ and $E_2^{1\times}$ allows us to remove the unstable correlation.
  }\label{tab:correlation}
  \vspace{-3mm}
\end{table}

\begin{table*}[t]
  \centering
  \begin{tabular}{lcccccccccc}
  \toprule
  &
  \multicolumn{2}{c}{\textsc{erm}} &
  \multicolumn{2}{c}{\textsc{dro}} &
  \multicolumn{2}{c}{\textsc{irm}} &
  \multicolumn{2}{c}{\textsc{rgm}} &
  \multicolumn{2}{c}{\textsc{pi (ours)}}
  \\
  \cmidrule(lr{0.5em}){2-3}
  \cmidrule(lr{0.5em}){4-5}
  \cmidrule(lr{0.5em}){6-7}
  \cmidrule(lr{0.5em}){8-9}
  \cmidrule(lr{0.5em}){10-11}
  & Worst & Avg
  & Worst & Avg
  & Worst & Avg
  & Worst & Avg
  & Worst & Avg
  \\
  \midrule
Kidney cancer        & 16.67 & 66.66 & \textbf{50.00} & 68.70 & 33.33 & 73.07 & 33.33 & 71.31 & \textbf{50.00} & \textbf{74.76} \\
\midrule
Adenocarcinoma       & 33.33 & 72.91 & 77.29 & 79.23 & 55.56 & 78.40 & 55.56 & 78.12 & \textbf{80.24} & \textbf{84.74} \\
\midrule
Lung cancer          & 44.44 & 74.76 & 62.50 & 74.58 & 38.89 & 74.28 & 50.00 & 74.75 & \textbf{70.31} & \textbf{78.85} \\
\midrule
Polyp syndrom        & 44.44 & 74.63 & \textbf{77.29} & 78.79 & 55.56 & 76.31 & 66.67 & 78.73 & 69.23 & \textbf{81.28} \\
\midrule
Hepatobiliary cancer & 44.44 & 73.01 & \textbf{ 60.00} & 73.89 & 55.56 & 77.19 & 55.56 & 76.22 & \textbf{60.00} & \textbf{78.94} \\
\midrule
Breast cancer        & 66.49 & 80.47 &  75.00 & 78.84 & 66.87 & 80.56 & 64.38 & 79.81 & \textbf{80.32} & \textbf{83.12} \\
\midrule
\midrule
Average$^*$          & 54.80 & 77.73 & 67.34 & 77.58 & 57.86 & 79.18 & 60.81 & 79.03 & \textbf{74.15} & \textbf{83.15} \\
  \bottomrule
  \end{tabular}
  \caption{
    Worst-group and average-group accuracy on ASK2ME text classification.
    We show the results for the worst 5 attributes (sorted based on ERM) and the given attribute \texttt{breast\_cancer}.
    Average is computed based on the performance across all attributes.
    See Appendix~\ref{app:results} for full results.
  }\label{tab:pubmed_small}
  \vspace{-4mm}
\end{table*}

\section{Results}\label{sec:results}

\subsection{Synthetic environments}
Table~\ref{tab:syn} summarizes the results on synthetic environments.
As we expected, 
since the signs of the unstable correlation are the same across the training environments,
both ERM and DRO exploit this information and fail to generalize when it changes in the testing environment.
While IRM and RGM are able to learn stable correlations when we use the testing environment for model selection,
their performance quickly drop to that of ERM when the validation data is drawn from the training environment, which also backs up the claim from \citet{gulrajani2020search}.

\begin{figure}[t]
  \centering
  \includegraphics[width=0.9\linewidth]{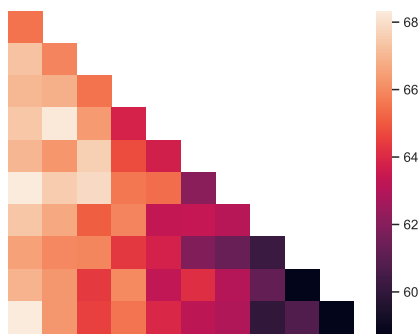}
  \caption{
    The ability of generalization changes as we vary the gap between the
    training environments.
    The x and y axes denote the probabilities that the injected artificial token
    agrees with the label.
    Heatmap corresponds to the testing accuracy for Beer Aroma.
  }\label{fig:ablation}
  \vspace{-4mm}
\end{figure}

Our algorithm obtains substantial gains across four tasks
It performs much more stable under different validation settings.
Specifically, comparing against the best baseline, our algorithm improves the
accuracy by 20.06\% when the validation set is drawn from the training
environment and 12.97\% when it is drawn from the testing environment.
Its performance closely matches the oracle model with only 2\% difference on
average.

\begin{figure}[t]
  \centering
  \includegraphics[width=\linewidth]{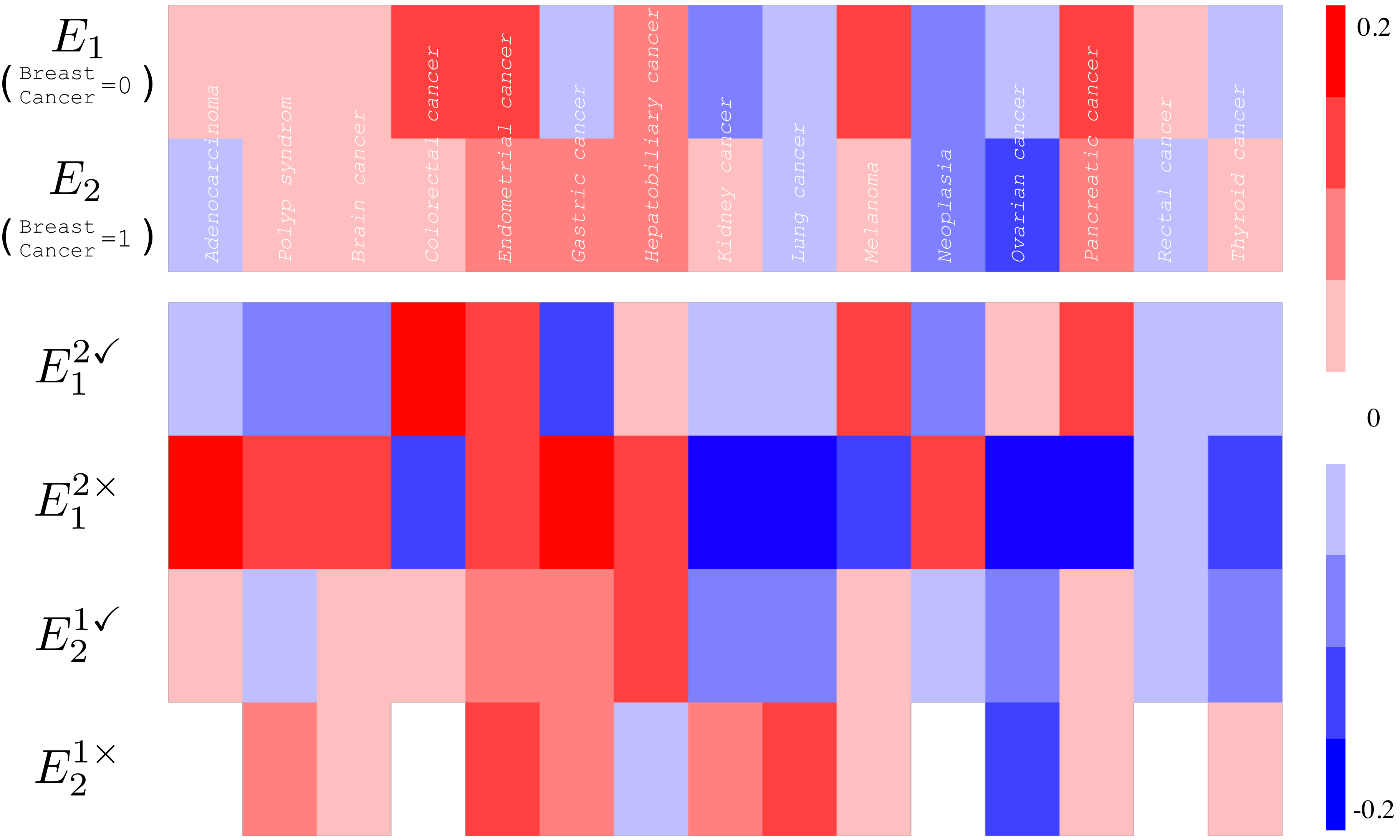}
  \caption{Visualization of the Pearson correlation coefficient between the
    label and the attribute on ASK2ME.
    Each column corresponds to a different attribute.
    We observe that correlations vary for inputs with different \texttt{breast\_cancer}
    value.
    Our algorithm utilizes this difference to create partitions with opposite
    correlations (red vs. blue) so that we can uncover an oracle distribution (different for each attribute) by interpolating these partitions.
  }\label{fig:pubmed}
  \vspace{-4mm}
\end{figure}

\textbf{\emph{Why does partitioning the training environments help?}}
To demystify the huge performance gain,
we quantitatively analyze the partitions created by our algorithm in
Table~\ref{tab:correlation}.\footnote{The partitions only depend on
  the training environments. It is independent to the choice of
  the validation data.}
We see that while the unstable correlation is positive in both training
environments,
it flips to negative in the set of wrong predictions, confirming our theoretical
analysis.
In order to perform well across all partitions,
our final classifier learns not to rely on the unstable features.

\begin{figure*}[t]
  \centering
  \includegraphics[width=\linewidth]{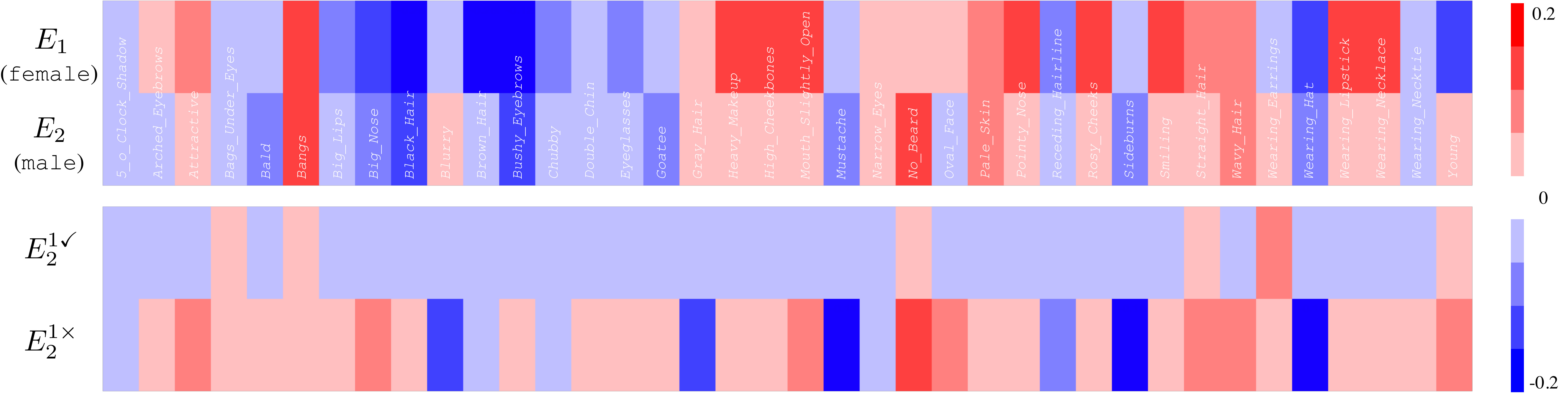}
  \vspace{-5mm}
  \caption{Visualization of the Pearson correlation coefficient between the
    label (hair color) and other attributes on CelebA.
    Each column corresponds to a different attribute.
    Due to the huge difference between the label marginals,
    $P_1(\texttt{blond})=0.24$ vs. $P_2(\texttt{blond})=0.02$,
    classifier $f_2$ predicts every example in environment $E_1$ as \texttt{dark}.
    The resulting partition, $E_1^{2\times}=\{\texttt{female, blond}\}$ and
    $E_1^{2\checkmark}=\{\texttt{female, dark}\}$, coincides with the human-defined groups in DRO.
    On the other hand, classifier $f_1$ is able to partition environment $E_2$ with
    opposite correlations (red vs. blue).
  }\label{fig:celeba}
\end{figure*}

\begin{table*}[t]
  \centering
  \begin{tabular}{lcccccccccc}
  \toprule
  &
  \multicolumn{2}{c}{\textsc{erm}} &
  \multicolumn{2}{c}{\textsc{dro}} &
  \multicolumn{2}{c}{\textsc{irm}} &
  \multicolumn{2}{c}{\textsc{rgm}} &
  \multicolumn{2}{c}{\textsc{pi (ours)}}
  \\
  \cmidrule(lr{0.5em}){2-3}
  \cmidrule(lr{0.5em}){4-5}
  \cmidrule(lr{0.5em}){6-7}
  \cmidrule(lr{0.5em}){8-9}
  \cmidrule(lr{0.5em}){10-11}
  & Worst & Avg
  & Worst & Avg
  & Worst & Avg
  & Worst & Avg
  & Worst & Avg
  \\
  \midrule
Goatee                & 0.00  & 68.26 & 0.00  & 70.08 & 84.80 & 90.83 & 91.50 & \textbf{95.66} & \textbf{91.63} & 94.59 \\
\midrule
Wearing\_Hat          & 7.69  & 70.39 & 46.15 & 82.28 & 46.15 & 77.34 & \textbf{61.54} & \textbf{86.26} & 53.85 & 84.56 \\
\midrule
Chubby                & 9.52  & 70.69 & 61.90 & 84.63 & \textbf{76.19} & 82.91 & 47.62 & 82.32 & 71.43 & \textbf{86.59} \\
\midrule
Wearing\_Necktie      & 25.00 & 74.52 & 90.00 & 91.56 & 80.00 & 84.31 & 35.00 & 79.32 & \textbf{91.44} & \textbf{92.53} \\
\midrule
Sideburns             & 38.46 & 77.84 & 84.62 & 90.69 & 76.92 & 84.14 & 76.92 & 89.72 & \textbf{91.35} & \textbf{93.75} \\
\midrule
Gender                & 46.67 & 80.14 & 85.56 & 90.87 & 74.44 & 83.93 & 70.00 & 87.73 & \textbf{90.56} & \textbf{91.52} \\
\midrule \midrule
Average$^*$       & 60.06 & 83.63 & 84.25 & 90.83 & 78.51 & 85.79 & 82.52 & 90.95 & \textbf{87.04} & \textbf{91.41} \\
  \bottomrule
  \end{tabular}
  \caption{
    Worst-group and average-group accuracy for hair color prediction on CelebA.
     We show the results for the worst 5 attributes (sorted based on ERM) and the given  attribute \texttt{gender}.
    Average is computed based on the performance across all attributes.
    See Appendix~\ref{app:results} for full results.
  }\label{tab:celeba_small}
    \vspace{-3mm}
\end{table*}

\textbf{\emph{Do we need different training environments?}}
We study the relation between the diversity of the training environments
and the performance of the classifier on the beer review dataset.
Specifically, we consider 45 different training environment pairs
where we vary the probability that the artificial token agrees with the label
from $0.80$ to $0.89$.
We observe that the classifier performs better as we reduce the amount of spurious correlations (moving up along the diagonal in Figure~\ref{fig:ablation}).
The classifier also generalizes better when we increase the gap between the two training environments (moving from right to left in Figure~\ref{fig:ablation}).
In fact, when the training environments share the same distribution, the
notion of stable correlation and unstable correlation is undefined.
There is no signal for the algorithm to distinguish between spurious features
and features that generalize.



\subsection{Natural environments}
Table~\ref{tab:pubmed_small} and ~\ref{tab:celeba_small} summarize
the results on using natural environments to reduce biases from attributes that
are unknown during both training and validation.
We observe that directly applying DRO over human-defined groups already surpasses IRM and RGM on the worst-case accuracy averaged across all attributes.
In addition,
for the given attribute (\texttt{breast\_cancer} and \texttt{gender}),
DRO achieves nearly 10\% more improvements over other baselines.
However, this robustness doesn't generalize equally towards other attributes.
By using the environment-specific classifier to create groups with contrasting unstable correlations,
our algorithm delivers marked performance improvement over DRO, 6.81\% on ASK2ME and 2.79\% on CelebA.

\textbf{\emph{How do we reduce bias from unknown attributes?}}
Figure~\ref{fig:pubmed} and \ref{fig:celeba} visualize the correlation between each attribute and
the label on ASK2ME and CelebA.
We observe that although the signs of the correlation can be the same
across the training environments, their magnitude may vary.
Our algorithm makes use of this difference to create partitions that have
opposite correlations for 13 (out of 15) attributes on ASK2ME and 22 (out of 38) attributes on CelebA.
These opposite correlations help the classifier to avoid using unstable features during training.

\section{Conclusion}\label{sec:conclusion}
In this paper, we propose a simple algorithm to learn correlations that are
stable across environments.
Specifically, we propose to use a classifier that is trained on one
environment to partition another environment.
By interpolating the distributions of its correct predictions and wrong
predictions, we can uncover an oracle distribution where the unstable
correlation vanishes.
Experimental results on synthetic environments and natural environments
validate that our algorithm is able to generate paritions with opposite unstable
correlations and reduce bias that are unknown during training.

\section*{Acknowledgement}
We thank all reviewers and the MIT NLP group for their thoughtful feedback.
Research was sponsored by the United States Air Force Research Laboratory and the United States Air Force Artificial Intelligence Accelerator and was accomplished under Cooperative Agreement Number FA8750-19-2-1000. The views and conclusions contained in this document are those of the authors and should not be interpreted as representing the official policies, either expressed or implied, of the United States Air Force or the U.S. Government. The U.S. Government is authorized to reproduce and distribute reprints for Government purposes notwithstanding any copyright notation herein.


\bibliography{main}
\bibliographystyle{icml2020}




\clearpage
\appendix

\section{A toy example}
\label{app:toy_verify}
We would like to show that the optimal solution that minimize the worst-case risk across $E_2^{1\checkmark}$ and $E_2^{1\times}$ is to predict $Y$ only using $X_1$.
Consider any classifier $f(Y \mid X_1, X_2)$ and its marginal
\[
f(Y\mid X_1) \propto f(X_1, X_2=0, Y) + f(X_1, X_2=1, Y).
\]
For any input $(x_1, x_2) \in E_2^{1\checkmark}$,
based on our construction,
the distribution $P_2^{1\checkmark}(X_1=x_1, X_2=x_2, Y)$ only has mass on one label value $y\in \{0, 1\}$.
Thus $P_2^{1\checkmark}(Y=y \mid X_1=x_1, X_2=x_2) = 1$.
We can then write the log risk of the classifier $f(Y \mid X_1, X_2)$ as
\[
-\log \frac{f(x_1, x_2, y)}{f(x_1, x_2, y) + f(x_1, x_2, 1-y)}.
\]
The log risk of the marginal classifier $f(Y \mid X_1)$ is defined as
\begin{align*}
-\log \Big(&(f(x_1, x_2, y) + f(x_1, 1-x_2, y)) \\
&/ (f(x_1, x_2, y) + f(x_1, 1-x_2, y) \\
&\quad + f(x_1, x_2, 1-y) + f(x_1, 1-x_2, 1-y))\Big).
\end{align*}
Now suppose $f(Y\mid X_1, X_2)$ achieves a lower risk than $f(Y\mid X_1)$.
This implies
\begin{align*}
&f(x_1, x_2, y) f(x_1, x_2, y) + f(x_1, x_2, 1-y) f(x_1, x_2, y)\\
&\qquad + f(x_1, x_2, y) f(x_1, 1-x_2, y)\\
&\qquad + f(x_1, x_2, 1-y) f(x_1, 1-x_2, y)\\
& < f(x_1, x_2, y) f(x_1, x_2, y) + f(x_1, x_2, y) f(x_1, x_2, 1-y)\\
&\qquad + f(x_1, x_2, y) f(x_1, 1-x_2, y)\\
&\qquad + f(x_1, x_2, y) f(x_1, 1-x_2, 1-y).
\end{align*}
Note that the first three terms on both side cancel out.
We have
\begin{align*}
&f(x_1, x_2, 1-y) f(x_1, 1-x_2, y) \\
&\qquad \qquad < f(x_1, x_2, y) f(x_1, 1-x_2, 1-y).
\end{align*}

Now let's consider an input $(x_1, 1-x_2) \in E_2^{1\times}$.
Based on our construction of the partitions,
we have $P_2^{1\checkmark}(x_1, x_2, y) = P_2^{1\times}(x_1, 1-x_2, y)$.
The log risk of the marginal classifier on $P_2^{1\times}$ is still the same, but the log risk of the classifier $f(Y \mid X_1, X_2)$ now becomes
\[
- \log \frac{f(x_1, 1-x_2, y)}{f(x_1, 1-x_2, y) + f(x_1, 1-x_2, 1-y)}.
\]
We claim that the log risk of $f(Y \mid X_1, X_2)$ is higher than $f(Y \mid X_1)$ on $P_2^{1\times}$.
Suppose for contradiction that the log risk of  $f(Y \mid X_1, X_2)$  is lower,
then we have
\begin{align*}
&f(x_1, 1-x_2, y) f(x_1, x_2, y) + f(x_1, 1-x_2, 1-y) f(x_1, x_2, y)\\
&\qquad + f(x_1, 1-x_2, y) f(x_1, 1-x_2, y)\\
&\qquad + f(x_1, 1-x_2, 1-y) f(x_1, 1-x_2, y)\\
& < f(x_1, 1-x_2, y) f(x_1, x_2, y) + f(x_1, 1-x_2, y) f(x_1, 1-x_2, y)\\
&\qquad + f(x_1, 1-x_2, y) f(x_1, x_2, 1-y)\\
&\qquad + f(x_1, 1-x_2, y) f(x_1, 1-x_2, 1-y).
\end{align*}
Canceling out the terms, we obtain
\begin{align*}
&f(x_1, 1-x_2, 1-y) f(x_1, x_2, y) \\
&\qquad \qquad < f(x_1, 1-x_2, y) f(x_1, x_2, 1-y).
\end{align*}
Contradiction!

Thus the marginal $f(Y\mid X_1)$ will always reach a better worst-group risk compare to the original classifier $f(Y \mid X_1, X_2)$.
As a result, the optimal classifier $f(Y \mid X_1, X_2)$ should satisfy $f(Y \mid X_1, X_2) = f(Y \mid X_1)$, i.e., it will only use $X_1$ to predict $Y$.
\section{Theoretical analysis}
\label{app:theory}

\addtocounter{prop}{-1}
\begin{prop}
  For a pair of environments $E_i$ and $E_j$,
  assuming that the classifier $f_i$ is able to learn the true conditional
  $P_i(Y\mid X_1, X_2)$,
  we can write the joint distribution $P_j$ of $E_j$ as the mixture of
  $P_j^{i \checkmark}$ and $P_j^{i \times}$:
  \[
    P_j(x_1, x_2, y) = \alpha_j^i P_j^{i \checkmark} (x_1, x_2, y) +
    (1-\alpha_j^i)  P_j^{i \times} (x_1, x_2, y),
  \]
  where $\alpha_j^i = \sum_{x_1, x_2, y} P_j(x_1, x_2, y)\cdot P_i(y \mid x_1, x_2)$ and
  \[
    \begin{aligned}
      P_j^{i\checkmark}(x_1, x_2, y) &\propto
      P_j(x_1, x_2, y)\cdot P_i(y \mid x_1, x_2),\\
      P_j^{i\times}(x_1, x_2, y) &\propto
      P_j(x_1, x_2, y)\cdot P_i(1-y \mid x_1, x_2).
    \end{aligned}
  \]
\end{prop}

\begin{proof}
For ease of notation, let $i=1$, $j=2$.
For an input $(x_1, x_2)$,
let's first consider the conditional probability $P_2^{1\times}(y \mid x_1, x_2)$  and $P_2^{1\checkmark}(y \mid x_1, x_2)$.
Since the input is in $E_2$, the probability that it has label $y$ is given by $P_2(y\mid x_1, x_2)$.
Since $f_1$ matches $P_1(y \mid x_1, x_2)$, the likelihood that the prediction is wrong is given by $P_1(1-y \mid x_1, x_2)$ and the likelihood that the prediction is correct is givn by $P_1(y \mid x_1, x_2)$.
Thus, we have
\[
\begin{aligned}
P_2^{1\times }(y \mid x_1, x_2) &=
\frac{ P_1(1-y \mid x_1, x_2) P_2(y \mid x_1, x_2)}{\sum_{y'}P_1(1-y' \mid x_1, x_2)  P_2(y' \mid x_1, x_2)},\\
P_2^{1\checkmark }(y \mid x_1, x_2) &=
\frac{ P_1(y \mid x_1, x_2) P_2(y \mid x_1, x_2)}{\sum_{y'}P_1(y' \mid x_1, x_2)  P_2(y' \mid x_1, x_2)}.
\end{aligned}
\]
Now let's think about the marginal of $(x_1, x_2)$ if it is in the set of mistakes $E_2^{1\times}$.
Again, since the input is in $E_2$, the probability that it exists is given by the marginal in $E_2$: $P_2(x_1, x_2)$.
This input has two possibilities to be partitioned into $E_2^{1\times}$: 1) the label is $y$ and $f_1$ predicts it as $1-y$; 2) the label is $1-y$ and $f_1$ predicts it as $y$.
Marginalizing over all $(x_1, x_2)$, we have
\[
\begin{aligned}
& P_2^{1\times} (x_1, x_2)\\
&= \frac{\frac{P_2(x_1, x_2) \sum_y P_1(1-y \mid x_1, x_2) P_2(y \mid x_1, x_2)}{\sum_y P_1(1-y \mid x_1, x_2) P_2(y \mid x_1, x_2) + P_1(y \mid x_1, x_2) P_2(y \mid x_1, x_2)}}{\sum_{x_1', x_2'}\frac{P_2(x_1', x_2') \sum_y P_1(1-y \mid x_1', x_2') P_2(y \mid x_1', x_2')}{\sum_y P_1(1-y \mid x_1', x_2') P_2(y \mid x_1', x_2') + P_1(y \mid x_1', x_2') P_2(y \mid x_1', x_2')}}\\
&=\frac{P_2(x_1, x_2) \sum_y P_1(1-y \mid x_1, x_2) P_2(y \mid x_1, x_2)}{\sum_{x_1', x_2'} P_2(x_1', x_2') \sum_y P_1(1-y \mid x_1', x_2') P_2(y \mid x_1', x_2')}
\end{aligned}
\]
Similarly, we have
\[\begin{aligned}
&P_2^{1\checkmark} (x_1, x_2)\\
&=\frac{P_2(x_1, x_2) \sum_y P_1(y \mid x_1, x_2) P_2(y \mid x_1, x_2)}{\sum_{x_1', x_2'} P_2(x_1', x_2') \sum_y P_1(y \mid x_1', x_2') P_2(y \mid x_1', x_2')}
\end{aligned}\]
Combining these all together using the Bayes' theorem, we have
\[
\begin{aligned}
&P_2^{1\times}(x_1, x_2, y)\\
&=\frac{ P_1(1-y \mid x_1, x_2) P_2(y \mid x_1, x_2) P_2(x_1, x_2)}{\sum_{x_1', x_2'} P_2(x_1', x_2') \sum_{y'} P_1(1-y' \mid x_1', x_2') P_2(y' \mid x_1', x_2')},\\
&=\frac{ P_1(1-y \mid x_1, x_2) P_2(x_1, x_2, y)}{\sum_{x_1', x_2', y'} P_2(x_1', x_2', y') P_1(1-y' \mid x_1', x_2')},\\
&\propto P_1(1-y \mid x_1, x_2) P_2(x_1, x_2, y),\\
&P_2^{1\checkmark}(x_1, x_2, y)\\
&=\frac{ P_1(y \mid x_1, x_2) P_2(y \mid x_1, x_2) P_2(x_1, x_2)}{\sum_{x_1', x_2'} P_2(x_1', x_2') \sum_{y'} P_1(y' \mid x_1', x_2') P_2(y' \mid x_1', x_2')},\\
&=\frac{ P_1(y \mid x_1, x_2) P_2(x_1, x_2, y)}{\sum_{x_1', x_2', y'} P_2(x_1', x_2', y') P_1(y' \mid x_1', x_2')},\\
&\propto P_1(y \mid x_1, x_2) P_2(x_1, x_2, y).
\end{aligned}
\]
Finally, it is straightforward to show that for $\alpha_2^1 = \sum_{x_1, x_2, y} P_2(x_1, x_2, y) P_1(y\mid x_1, x_2)$,
we have
\[
\begin{aligned}
&\alpha_2^1 P_2^{1\checkmark} (x_1, x_2, y) + (1-\alpha_2^1) P_2^{1\times}\\
&= P_1(y \mid x_1, x_2) P_2(x_1, x_2, y) \\
&\qquad\qquad\qquad + P_1(1-y \mid x_1, x_2) P_2(x_1, x_2, y)\\
&= P_2(x_1, x_2, y).
\end{aligned}
\]
\end{proof}

From now on, we assume that the marginal distribution of $Y$ is uniform in all joint distributions, i.e., $f_i$ performs equally well on different labels.
\addtocounter{thm}{-2}
\begin{thm}
  Suppose $X_2$ is independent of $X_1$ given $Y$.
  For any environment pair $E_i$ and $E_j$,
  if $\sum_y P_i(x_2 \mid y) = \sum_y P_j(x_2 \mid y)$ for any $x_2$,
  then $\mathrm{Cov}(X_2, Y; P_i) > \mathrm{Cov}(X_2, Y; P_j)$ implies
  $\mathrm{Cov}(X_2, Y; P_j^{i\times}) < 0$ and
  $\mathrm{Cov}(X_2, Y; P_i^{j\times}) > 0$.
\end{thm}
\begin{proof}
By definition, we have
\[\begin{aligned}
&\mathrm{Cov}(X_2, Y; P_j^{i\times})\\
&= \mathbb{E}[X_2 Y; P_j^{i\times}] - \mathbb{E}[X_2; P_j^{i\times}]\, \mathbb{E}[Y; P_j^{i\times}]\\
&= \sum_{x_1, x_2} x_2 P_j^{i\times}(x_1, x_2, 1) \\
&\quad - \sum_{x_1, x_2, y} x_2 P_j^{i\times}(x_1, x_2, y)
\sum_{x_1, x_2} P_j^{i\times}(x_1, x_2, 1)\\
&= \sum_{x_1, x_2, x_1', x_2', y'} x_2 P_j^{i\times}(x_1, x_2, 1) P_j^{i\times}(x_1', x_2', y') \\
&\quad - \sum_{x_1, x_2, y, x_1', x_2'} x_2 P_j^{i\times}(x_1, x_2, y)
 P_j^{i\times}(x_1', x_2', 1)\\
\end{aligned}
\]
Expanding the distributions of $P_j^{i\times}$,
it suffices to show that
\[
\begin{aligned}
&\sum_{x_1, x_2, x_1', x_2', y'} \Big( x_2 P_j(x_1, x_2, 1) P_i(0 \mid x_1, x_2)\\[-10pt]
&\qquad\qquad\qquad\quad P_j(x_1', x_2', y') P_i(1-y' \mid x_1', x_2')\Big) \\
&< \sum_{x_1, x_2, y, x_1', x_2'}\Big( x_2 P_j(x_1, x_2, y) P_i(1-y \mid x_1, x_2)\\[-10pt]
&\qquad\qquad\qquad\quad P_j(x_1', x_2', 1) P_i(0 \mid x_1', x_2')\Big) \\
\end{aligned}
\]
Note that when $y=y'=1$, two terms cancel out. Thus we need to show
\[
\begin{aligned}
&\sum_{x_1, x_2, x_1', x_2'} \Big( x_2 P_j(x_1, x_2, 1) P_i(0 \mid x_1, x_2)\\[-10pt]
&\qquad\qquad\qquad\quad P_j(x_1', x_2', 0) P_i(1 \mid x_1', x_2')\Big) \\
&< \sum_{x_1, x_2, x_1', x_2'}\Big( x_2 P_j(x_1, x_2, 0) P_i(1 \mid x_1, x_2)\\[-10pt]
&\qquad\qquad\qquad\quad P_j(x_1', x_2', 1) P_i(0 \mid x_1', x_2')\Big) \\
\end{aligned}
\]
Based on the assumption that the marginal distribution in $E_j^{i \times}$ is uniform, we have
\[\begin{aligned}
&\sum_{x_1', x_2'} P_j(x_1', x_2', 0) P_i(1 \mid x_1', x_2')\\
&\doteq \sum_{x_1', x_2'} P_j(x_1', x_2', 1) P_i(0 \mid x_1', x_2').
\end{aligned}\]
Thus we can simplify our goal as
\[
\begin{aligned}
&\sum_{x_1, x_2}  x_2 P_j(x_1, x_2, 1) P_i(0 \mid x_1, x_2)\\
&< \sum_{x_1, x_2} x_2 P_j(x_1, x_2, 0) P_i(1 \mid x_1, x_2)\\
\end{aligned}
\]
Similarly, we can simplify the condition $\mathrm{Cov}(X_2, Y; P_i) > \mathrm{Cov}(X_2, Y; P_j)$ as
\[
\begin{aligned}
&\sum_{x1, x_2} x_2 (P_j(x_1, x_2, 1) - P_i(x_1, x_2, 1))\\
&<\sum_{x1, x_2} x_2 (P_j(x_1, x_2, 0) - P_i(x_1, x_2, 0))
\end{aligned}
\]
Since $x_2$ is independent of $x_1$ given $y$, we have
\[
\begin{aligned}
&\sum_{x1, x_2} x_2 (P_j(x_1, y=1) P_j(x_2 \mid y=1) -\\[-10pt]
&\qquad\qquad\qquad P_i(x_1, y=1) P_i(x_2 \mid y=1) )\\
&<\sum_{x1, x_2} x_2 (P_j(x_1, y=0) P_j(x_2 \mid y=0) -\\[-10pt]
&\qquad\qquad\qquad P_i(x_1, y=0) P_i(x_2 \mid y=0))
\end{aligned}
\]
Since $x_1$ is the stable feature and the label marginal is the same across environments, we have $P_j(x_1, y=1) = P_i(x_1, y=1)$ and  $P_j(x_1, y=0) = P_i(x_1, y=0)$. This implies
\[
\begin{aligned}
&\sum_{x_1} P_j(x_1, y=1) \sum_{x_2} x_2 (P_j(x_2 \mid y=1) -P_i(x_2 \mid y=1) )\\
&<\sum_{x_1} P_j(x_1, y=0) \sum_{x_2} x_2 (P_j(x_2 \mid y=0) -P_i(x_2 \mid y=0) )\\
\end{aligned}
\]
Again, by uniform label marginals, we have
\[
\begin{aligned}
&\sum_{x_2} x_2 (P_j(x_2 \mid y=1) -P_i(x_2 \mid y=1))\\
&<\sum_{x_2} x_2 (P_j(x_2 \mid y=0) -P_i(x_2 \mid y=0).
\end{aligned}
\]
For binary $x_2 \in \{ 0,1 \}$, this implies
$P_j(x_2 = 1 \mid y = 1) + P_i(x_2=1 \mid y = 0) < P_j(x_2=1\mid y=0) + P_i(x_2=1 \mid y = 1).$
Since $P_j(x_2 \mid y=1) + P_j(x_2 \mid y=0) = P_i(x_2 \mid y=1) + P_i(x_2 \mid y=0)$, we have

\begin{equation}\label{eq:1}
\begin{aligned}
&P_j(x_2\mid y=1) P_j (x_2 \mid y=0)\\
&\qquad < P_j(x_2\mid y=0) P_j (x_2 \mid y=1).
\end{aligned}
\end{equation}

We can expand our goal in the same way:
\[
\begin{aligned}
&\sum_{x_1, x_2}  x_2 P_j(x_1, x_2, 1) P_i(0 \mid x_1, x_2)\\
&=\sum_{x_1, x_2} \Big(x_2 P_j (x_1, y=1) P_i(x_1, y=0) \\
&\qquad\qquad P_j(x_2 \mid y=1)  P_i(x_2 \mid y=0)\Big)/P_i(x_1, x_2) \\
&=\sum_{x_1}P_j (x_1, y=1) P_i(x_1, y=0)  \\
&\qquad \cdot \sum_{x_2}\frac{x_2 P_j(x_2 \mid y=1) P_i(x_2 \mid y=0)}{P_i(x_1, x_2)}
\end{aligned}
\]
\[
\begin{aligned}
&\sum_{x_1, x_2}  x_2 P_j(x_1, x_2, 0) P_i(1 \mid x_1, x_2)\\
&=\sum_{x_1}P_j (x_1, y=0) P_i(x_1, y=1)  \\
&\qquad \cdot 
\sum_{x_2}\frac{x_2 P_j(x_2 \mid y=0) P_i(x_2 \mid y=1)}{P_i(x_1, x_2)},
\end{aligned}
\]
Plug in Eq (\ref{eq:1}) and we complete the proof.
The other inequality follows by symmetry.
\end{proof}

\paragraph{Extension to multi-class classification:}
In Theorem~\ref{thm:1}, we focus on binary classification for simplicity. For multi-class classification, we can convert it into a binary problem by defining $Y_c$ as a binary indicator of whether class $c$ is present or absent. Our strong empirical performance on MNIST (10-class classification) also confirms that our results generalize to the multi-class setting.

\begin{thm}
  \label{thm:2}
  For any environment pair $E_i$ and $E_j$,
  $\mathrm{Cov}(X_2, Y; P_i) > \mathrm{Cov}(X_2, Y; P_j)$ implies
  \[
    \begin{aligned}
& \mathrm{Cov}(X_2, Y; P_j^{i\times})\\
&< \frac{1-\alpha_j^i}{\alpha_{i}^{i}} \mathrm{Cov}(X_2, Y; P_i^{i\checkmark}) - 
\frac{1-\alpha_j^i}{\alpha_j^i} \mathrm{Cov}(X_2, Y; P_j^{i\checkmark})\\
& \mathrm{Cov}(X_2, Y; P_i^{j\times})\\
&> \frac{1-\alpha_i^j}{\alpha_j^j} \mathrm{Cov}(X_2, Y; P_j^{j\checkmark}) - 
\frac{1-\alpha_i^j}{\alpha_i^j}\mathrm{Cov}(X_2, Y; P_i^{j\checkmark})
    \end{aligned}
  \]
  where $P_i^{i\checkmark}$ is the distribution of the correct predictions when
  applying $f_i$ on $E_i$.
\end{thm}
\begin{proof}
From the proof in Theorem~\ref{thm:1}, we can write the condition $\mathrm{Cov}(X_2, Y; P_i) > \mathrm{Cov}(X_2, Y; P_j)$ as
\[
\begin{aligned}
&\sum_{x1, x_2} x_2 (P_j(x_1, x_2, 1) - P_i(x_1, x_2, 1))\\
&<\sum_{x1, x_2} x_2 (P_j(x_1, x_2, 0) - P_i(x_1, x_2, 0))
\end{aligned}
\]
Using $ P_i(0 \mid x_1, x_2) + P_i(1\mid x_1, x_2) = 1$,
\[
\begin{aligned}
&\sum_{x1, x_2} x_2 (P_j(x_1, x_2, 1) - P_i(x_1, x_2, 1))\\[-10pt]
&\qquad \qquad ( P_i(0 \mid x_1, x_2) + P_i(1\mid x_1, x_2))\\
&<\sum_{x1, x_2} x_2 (P_j(x_1, x_2, 0)- P_i(x_1, x_2, 0)) \\[-10pt]
&\qquad \qquad ( P_i(0 \mid x_1, x_2) + P_i(1\mid x_1, x_2))
\end{aligned}
\]
Since $P_i(x_1, x_2, 1) P_i(0 \mid x_1, x_2) $ and  $P_i(x_1, x_2, 0) P_i(1 \mid x_1, x_2) $ cancel out with each other. We have
\[
\begin{aligned}
&\sum_{x1, x_2} x_2 (P_j(x_1, x_2, 1) P_i(0\mid x_1, x_2) - \\[-10pt]
&\qquad \qquad\qquad  P_j(x_1, x_2, 0) P_i(1\mid x_1, x_2))\\
&<\sum_{x1, x_2} x_2 (P_i(x_1, x_2, 1) P_i(1 \mid x_1, x_2) - \\[-10pt]
&\qquad \qquad\qquad  P_i(x_1, x_2, 0) P_i(0 \mid x_1, x_2))\\
&\quad - \sum_{x1, x_2} x_2 (P_j(x_1, x_2, 1) P_i(1 \mid x_1, x_2) - \\[-10pt]
&\qquad \qquad\qquad  P_j(x_1, x_2, 0) P_i(0 \mid x_1, x_2))\\
\end{aligned}
\]
From the derivations in Theorem~\ref{thm:1}, we know that
\[\begin{aligned}
&\frac{1}{2 (1-\alpha_{j}^i)}  \mathrm{Cov}(X_2, Y; P_j^{i\times})  \\
&= \sum_{x1, x_2} x_2 \Big(P_j(x_1, x_2, 1) P_i(0\mid x_1, x_2) -\\[-10pt]
& \qquad \qquad \qquad P_j(x_1, x_2, 0) P_i(1\mid x_1, x_2)\Big)\\
&\frac{1}{2 \alpha_{j}^i}  \mathrm{Cov}(X_2, Y; P_j^{i\checkmark})  \\
&= \sum_{x1, x_2} x_2 \Big(P_j(x_1, x_2, 1) P_i(1\mid x_1, x_2)  \\[-10pt]
&\qquad \qquad \qquad P_j(x_1, x_2, 0) P_i(0\mid x_1, x_2)\Big)\\
&\frac{1}{2\alpha_{i}^i}  \mathrm{Cov}(X_2, Y; P_i^{i\checkmark}) \\
&= \sum_{x1, x_2} x_2 \Big(P_i(x_1, x_2, 1) P_i(1\mid x_1, x_2) -\\[-10pt]
&\qquad \qquad \qquad P_i(x_1, x_2, 0) P_i(0\mid x_1, x_2)\Big).
\end{aligned}\]
Combining these, we have
\[\begin{aligned}
& \mathrm{Cov}(X_2, Y; P_j^{i\times})\\
&< \frac{1-\alpha_j^i}{\alpha_{i}^{i}} \mathrm{Cov}(X_2, Y; P_i^{i\checkmark}) - 
\frac{1-\alpha_j^i}{\alpha_j^i} \mathrm{Cov}(X_2, Y; P_j^{i\checkmark})
\end{aligned}\]
Similarly, by using $ P_j(0 \mid x_1, x_2) + P_j(1\mid x_1, x_2) = 1$,
we can get
\[\begin{aligned}
& \mathrm{Cov}(X_2, Y; P_i^{j\times})\\
&> \frac{1-\alpha_i^j}{\alpha_j^j} \mathrm{Cov}(X_2, Y; P_j^{j\checkmark}) - 
\frac{1-\alpha_i^j}{\alpha_i^j}\mathrm{Cov}(X_2, Y; P_i^{j\checkmark})
\end{aligned}\]
\end{proof}

\section{Experimental Setup}
\subsection{Datasets and Models}
\label{app:dataset}
\subsubsection{MNIST}
\paragraph{Data}
We use the official train-test split of MNIST. Training environments are constructed from training split, with 14995 examples per environment. Validation data and testing data is constructed based on the testing split, with 2497 examples each. Following \citet{arjovsky2019invariant}, We convert each grey scale image into a $10\times 28 \times 28$ tensor, where the first dimension corresponds to the spurious color feature.

\paragraph{Model:}
The input image is passed to a CNN with 2 convolution layers and 2 fully connected layers.
We use the architecture from PyTorch's MNIST example\footnote{https://github.com/pytorch/examples/blob/master/mnist/main.py}.

\subsubsection{Beer Review}
\paragraph{Data}
We use the data processed by ~\citet{lei2016rationalizing}.
Reviews shorter than 10 tokens or longer than 300 tokens are filtered out.
For each aspect, we sample training/validation/testing data randomly from the dataset and maintain the marginal distribution of the label to be uniform.
Each training environment contains 4998 examples.
The validation data contains 4998 examples and the testing data contains 5000 examples.
The vocabulary sizes for the three aspects (look, aroma, palate) are: 10218, 10154 and 10086.
The processed data will be publicly available.

\paragraph{Model}
We use a standard CNN text classifier~\cite{kim-2014-convolutional}.
Each input is first encoded by pre-trained FastText embeddings~\cite{mikolov2018advances}.
Then it is passed into a 1D convolution layer followed by max pooling and ReLU activation.
The convolution layer uses filter size $3, 4, 5$.
Finally we attach a linear layer with Softmax to predict the label.

\subsubsection{CelebA}
\paragraph{Data} We use the official train/val/test split of CelebA~\cite{liu2015faceattributes}.
The training environment \{\texttt{female}\} contains 94509 examples and the training environment \{\texttt{male}\} contains 68261 examples. The validation set has $19867$ examples and the test set has 19962 examples.

\paragraph{Model}
We use the Pytorch torchvision implementation of the ResNet50 model, starting from pretrained weights.
We re-initalize the final layer to predict the target attribute \texttt{hair color}.

\subsubsection{ASK2ME}
\paragraph{Data}
Since the original data doesn't have a standard train/val/test split,
we randomly split the data and use 50\% for training, 20\% for validation, 30\%for testing.
There are 2227 examples in the training environment \{\texttt{breast\_cancer=0}\}, 1394 examples in the training environment \{\texttt{breast\_cancer=1}\}.
The validation set contains 1448 examples and the test set contains 2173 examples.
The vocabulary size is 16310.
The processed data will be publicly available.

\paragraph{Model}
The model architecture is the same as the one for Beer review.

\subsection{Implementation details}
\label{app:details}
\paragraph{For all methods:}
We use batch size $50$ and evaluate the validation performance every $100$ batch.
We apply early stopping once the validation performance hasn't improved in the past 20 evaluations.
We use Adam~\cite{kingma2014adam} to optimize the parameters and tune the learning rate $\in\{10^{-3}, 10^{-4}, 10^{-5}\}$.
For simplicity, we train all methods without data augmentation.
Following \citet{sagawa2019distributionally}, we apply strong regularizations to avoid over-fitting.
Specifically, we tune the dropout rate $\in\{0.1, 0.3, 0.5\}$ for text classification datasets (Beer review and ASK2ME) and tune the weight decay parameters $\in\{10^{-0}, 10^{-1}, 10^{-2}, 10^{-3}\}$ for image datasets (MNIST and CelebA).

\textbf{DRO} and \textbf{Ours}
We directly optimize the $\min-\max$ objective.
Specifically, at each step, we sample a batch of example from each group, and minimize the worst-group loss.
We found the training process to be pretty stable when using the Adam optimizer. On CelebA, we are able to match the performance reported by \citet{sagawa2019distributionally}.

\textbf{IRM}
We implement the gradient penalty based on the official implementation of IRM\footnote{https://github.com/facebookresearch/InvariantRiskMinimization}. The gradient penalty is applied to the last hidden layer of the network. We tune the weight of the penalty term $\in\{10^{-2},\, 10^{-1},\, 10^{0},\, 10^{1},\, 10^{2},\, 10^{3},\, 10^{4}\}$ and the annealing iterations $\in\{10, 10^2, 10^3\}$.

\textbf{RGM}
For the per-environment classifier in RGM, we use a MLP with one hidden layer. This MLP takes the last layer of the model as input and predicts the label. Similar to IRM, we tune the weight of the regret $\in\{10^{-2},\, 10^{-1},\, 10^{0},\, 10^{1},\, 10^{2},\, 10^{3},\, 10^{4}\}$ and the annealing iterations $\in\{10, 10^2, 10^3\}$.

\begin{table}[t]
\begin{tabular}{lllll}
\toprule
    & TIME          & Train & Val   & Test  \\
    \midrule
ERM    & 2 MIN 58 SEC & 83.61 & 81.21 & 15.65 \\ \midrule
IRM    & 3 MIN 37 SEC  & 83.42 & 80.41 & 12.89 \\ \midrule
RGM    & 3 MIN 7 SEC   & 82.60 & 81.41 & 13.97 \\ \midrule
DRO    & 17 MIN 19 SEC & 79.44 & 80.65 & 16.05 \\ \midrule
OURS   & 11 MIN 58 SEC & 65.04 & 71.16 & 71.56 \\
\midrule
ORACLE & 14 MIN 31 SEC & 68.96 & 72.28 & 70.04 \\
\bottomrule
\end{tabular}
\caption{Running time and model performance on MNIST. Here the validation data is sampled from the training environments. Our algorithm requires training additional environment-specific classifiers. However, it converges faster than DRO in the third stage (50 epochs vs. 72 epochs) and generalizes much better.}
\label{tab:time_mnist}
\end{table}

\begin{table}[t]
\begin{tabular}{lllll}
\toprule
    & TIME          & Train & Val   & Test  \\
    \midrule
ERM & 3 MIN 35 SEC  & 99.44 & 66.01 & 59.04 \\
    \midrule
IRM & 3 MIN 21 SEC  & 98.70 & 63.10 & 57.85 \\
    \midrule
RGM & 5 MIN 36 SEC  & 99.78 & 64.07 & 59.99 \\
    \midrule
DRO & 16 MIN 40 SEC & 86.77 & 77.66 & 67.34 \\
    \midrule
PI (Ours) & 18 MIN        & 97.09 & 78.64 & 74.14 \\
    \bottomrule
\end{tabular}
\caption{Running time and model performance on ASK2ME. Here the validation accuracy is computed based on the \texttt{breast\_cancer} attribute. The test accuracy is the average worst-group accuracy across all 17 attributes. Our algorithm's running time is similar to DRO.}
\label{tab:time_pubmed}
\end{table}

\subsection{Computing Infrastructure and Running Time Analysis}
We have used the following graphics cards for our experiments: Tesla V100-32GB, GeForce RTX 2080 Ti and A100-40G.

We conducted our running time analysis on MNIST and ASK2ME using GeForce RTX 2080 Ti.
Table~\ref{tab:time_mnist} and \ref{tab:time_pubmed} shows the results.
We observe that due to the direct optimization of the $\min \max$ objective, the running time of DRO, PI and Oracle is roughly $4$ times comparing to other methods (proportional to the number of groups).
Also, while our model needs to train additional environment-specific classifiers (comparing to DRO), its running time is very similar to DRO across the two datasets.
We believe by using the online learning algorithm proposed by ~\citet{sagawa2019distributionally}, we can further reduce the running time of our algorithm.

\section{Additional results}
\label{app:results}

\begin{figure}[h]
    \centering
    \includegraphics[width=\linewidth]{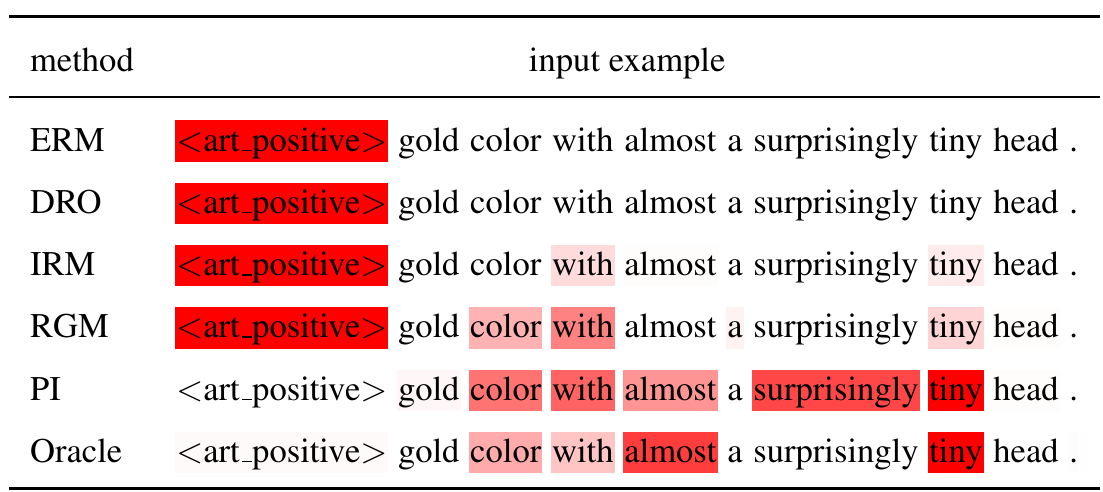}
    \vspace{-1em}
    \caption{Visualizing word importance on Beer Look. Only \texttt{PI} and \texttt{Oracle} ignore the artificial token and correctly predict the input as negative. We will add more examples in the update.}
    \label{fig:visual}
\end{figure}

\emph{\textbf{What features does \textsc{pi} look at?}}
To understand what features different methods rely on, we plot the word importance on Beer Look in Figure~\ref{fig:visual}. For the given input example, we evaluate the prediction change as we mask out each input token. We observe that only PI and Oracle ignore the spurious feature and predict the label correctly. Comparing to ERM, IRM and RGM focus more on the causal feature such as `tiny'. However, they still heavily rely on the spurious feature.

\begin{figure*}[t]
  \centering
  \includegraphics[width=\linewidth]{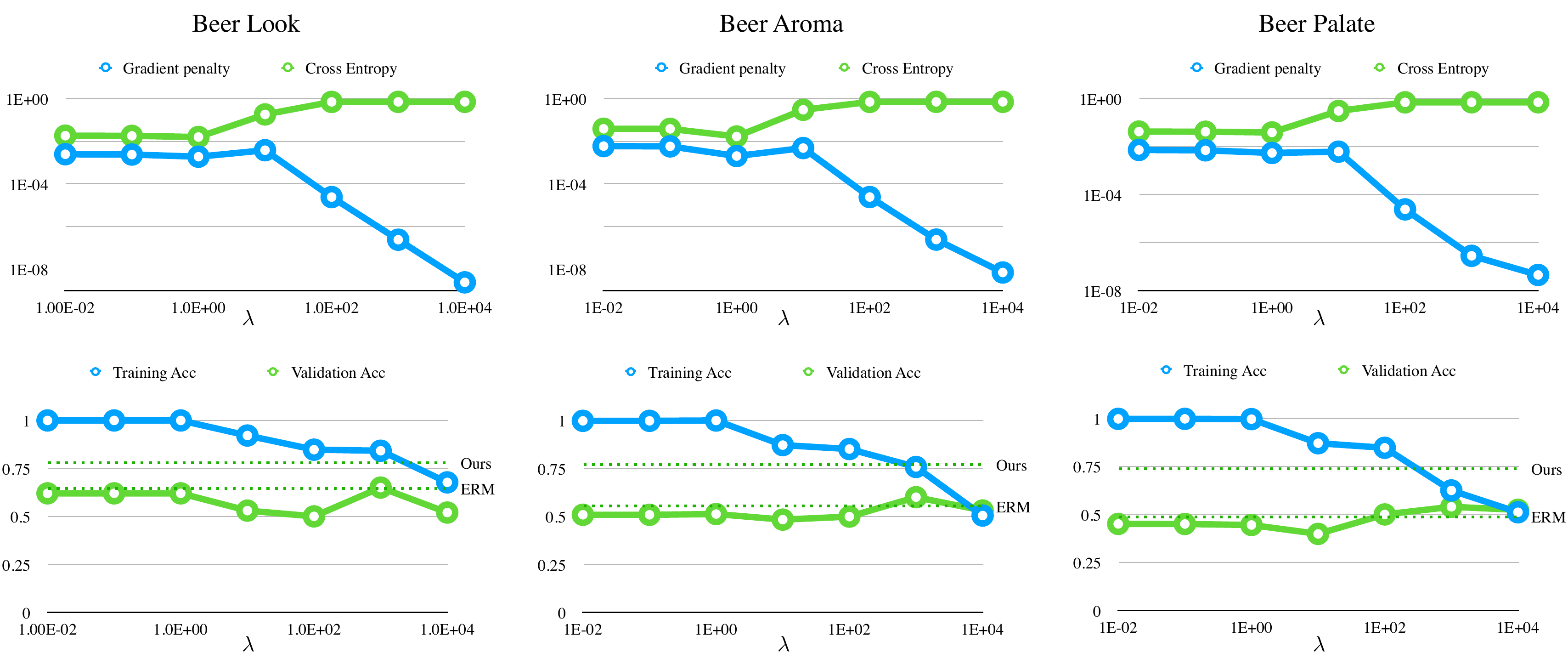}
  \caption{Performance of IRM as we adjust the weight of the gradient penalty. We observe that while the gradient penalty term is always orders of magnitude smaller than the cross entropy loss, the model is still able to overfit the unstable correlations in the training environments. As we further increase the penalty, the training \& validation performance quickly drop to that of ERM.}\label{fig:irm}
\end{figure*}

\begin{table*}[t]
  \centering
  \begin{tabular}{lcccccccccc}
  \toprule
  &
  \multicolumn{2}{c}{ERM} &
  \multicolumn{2}{c}{DRO} &
  \multicolumn{2}{c}{IRM} &
  \multicolumn{2}{c}{RGM} &
  \multicolumn{2}{c}{Ours}
  \\
  \cmidrule(lr{0.5em}){2-3}
  \cmidrule(lr{0.5em}){4-5}
  \cmidrule(lr{0.5em}){6-7}
  \cmidrule(lr{0.5em}){8-9}
  \cmidrule(lr{0.5em}){10-11}
  Accuracy
  & Worst & Avg
  & Worst & Avg
  & Worst & Avg
  & Worst & Avg
  & Worst & Avg
  \\
  \midrule
Adenocarcinoma       & 33.33 & 72.91 & 77.29 & 79.23 & 55.56 & 78.40 & 55.56 & 78.12 & 80.24 & 84.74 \\
Polyp syndrom        & 44.44 & 74.63 & 77.29 & 78.79 & 55.56 & 76.31 & 66.67 & 78.73 & 69.23 & 81.28 \\
Brain cancer         & 55.56 & 78.51 & 77.14 & 78.09 & 55.56 & 78.59 & 67.55 & 82.33 & 79.94 & 87.95 \\
Breast cancer        & 66.49 & 80.47 & 75.00 & 78.84 & 66.87 & 80.56 & 64.38 & 79.81 & 80.32 & 83.12 \\
Colorectal cancer    & 66.54 & 80.50 & 69.31 & 77.94 & 64.96 & 81.28 & 66.93 & 80.33 & 76.24 & 81.71 \\
Endometrial cancer   & 66.98 & 80.60 & 76.19 & 80.21 & 66.03 & 82.60 & 66.98 & 81.77 & 80.32 & 83.26 \\
Gastric cancer       & 62.96 & 79.94 & 76.95 & 81.65 & 62.96 & 80.03 & 59.26 & 78.87 & 79.44 & 85.92 \\
Hepatobiliary cancer & 44.44 & 73.01 & 60.00 & 73.89 & 55.56 & 77.19 & 55.56 & 76.22 & 60.00 & 78.94 \\
Kidney cancer        & 16.67 & 66.66 & 50.00 & 68.70 & 33.33 & 73.07 & 33.33 & 71.31 & 50.00 & 74.76 \\
Lung cancer          & 44.44 & 74.76 & 62.50 & 74.58 & 38.89 & 74.28 & 50.00 & 74.75 & 70.31 & 78.85 \\
Melanoma             & 66.67 & 80.55 & 66.67 & 78.87 & 66.67 & 83.32 & 66.67 & 79.69 & 80.06 & 86.67 \\
Neoplasia            & 50.00 & 75.98 & 33.33 & 69.10 & 33.33 & 71.97 & 50.00 & 75.18 & 70.00 & 80.06 \\
Ovarian cancer       & 65.31 & 80.16 & 77.20 & 79.30 & 66.80 & 80.64 & 66.33 & 79.53 & 73.47 & 82.76 \\
Pancreatic cancer    & 67.18 & 80.93 & 75.82 & 78.74 & 63.64 & 79.69 & 63.64 & 79.67 & 80.06 & 84.31 \\
Prostate cancer      & 63.96 & 85.77 & 51.04 & 77.48 & 64.29 & 85.21 & 65.58 & 83.92 & 78.90 & 86.75 \\
Rectal cancer        & 66.67 & 78.78 & 64.10 & 80.37 & 66.67 & 78.86 & 67.54 & 80.80 & 71.79 & 84.59 \\
Thyroid cancer       & 50.00 & 77.18 & 75.00 & 83.06 & 66.86 & 84.05 & 67.73 & 82.56 & 80.23 & 87.85 \\
\midrule
Average              & 54.80 & 77.73 & 67.34 & 77.58 & 57.86 & 79.18 & 60.81 & 79.03 & 74.15 & 83.15 \\
  \bottomrule
  \end{tabular}
  \caption{
    Worst-group and average-group accuracy across 17 attributes on ASK2ME.
  }\label{tab:pubmed_full}
\end{table*}

\begin{table*}[t]
  \centering
  \begin{tabular}{lcccccccccc}
  \toprule
  &
  \multicolumn{2}{c}{ERM} &
  \multicolumn{2}{c}{DRO} &
  \multicolumn{2}{c}{IRM} &
  \multicolumn{2}{c}{RGM} &
  \multicolumn{2}{c}{Ours}
  \\
  \cmidrule(lr{0.5em}){2-3}
  \cmidrule(lr{0.5em}){4-5}
  \cmidrule(lr{0.5em}){6-7}
  \cmidrule(lr{0.5em}){8-9}
  \cmidrule(lr{0.5em}){10-11}
  & Worst & Avg
  & Worst & Avg
  & Worst & Avg
  & Worst & Avg
  & Worst & Avg
  \\
  \midrule
5\_o\_Clock\_Shadow   & 53.33 & 81.51 & 90.00 & 92.05 & 80.00 & 85.93 & 66.67 & 87.22 & 83.33 & 89.65 \\
Arched\_Eyebrows      & 72.14 & 87.16 & 90.42 & 92.68 & 84.43 & 87.85 & 88.95 & 92.60 & 90.56 & 92.31 \\
Attractive            & 67.21 & 85.84 & 90.77 & 92.22 & 82.57 & 87.28 & 86.64 & 91.94 & 89.98 & 91.86 \\
Bags\_Under\_Eyes     & 72.46 & 86.16 & 90.59 & 92.04 & 81.34 & 86.84 & 88.52 & 92.13 & 89.12 & 92.12 \\
Bald                  & 75.98 & 91.23 & 91.73 & 93.04 & 71.39 & 82.21 & 91.50 & 94.81 & 91.68 & 93.42 \\
Bangs                 & 73.85 & 87.80 & 90.84 & 92.91 & 81.70 & 87.38 & 88.05 & 92.21 & 90.24 & 92.33 \\
Big\_Lips             & 73.46 & 87.14 & 90.59 & 92.55 & 84.16 & 87.87 & 89.54 & 92.65 & 90.52 & 92.17 \\
Big\_Nose             & 71.43 & 86.00 & 91.58 & 92.97 & 84.99 & 88.43 & 91.22 & 93.78 & 91.36 & 92.87 \\
Black\_Hair           & 75.98 & 90.91 & 89.62 & 93.77 & 78.63 & 89.16 & 90.66 & 94.02 & 88.10 & 93.30 \\
Blurry                & 51.23 & 81.06 & 86.56 & 90.14 & 79.36 & 85.73 & 79.01 & 89.71 & 85.61 & 89.39 \\
Brown\_Hair           & 43.68 & 79.17 & 64.37 & 85.74 & 78.16 & 83.39 & 72.41 & 87.30 & 59.77 & 83.83 \\
Bushy\_Eyebrows       & 72.73 & 86.52 & 72.73 & 88.83 & 81.82 & 87.51 & 81.82 & 91.26 & 81.82 & 90.77 \\
Chubby                & 9.52  & 70.69 & 61.90 & 84.63 & 76.19 & 82.91 & 47.62 & 82.32 & 71.43 & 86.59 \\
Double\_Chin          & 50.00 & 80.73 & 90.66 & 91.76 & 78.52 & 86.35 & 91.50 & 92.74 & 90.21 & 92.45 \\
Eyeglasses            & 58.06 & 82.83 & 90.32 & 92.02 & 80.44 & 85.71 & 77.42 & 89.34 & 88.71 & 91.17 \\
Goatee                & 0.00  & 68.26 & 0.00  & 70.08 & 84.80 & 90.83 & 91.50 & 95.66 & 91.63 & 94.59 \\
Gray\_Hair            & 60.71 & 82.53 & 69.08 & 87.73 & 42.60 & 76.20 & 85.71 & 89.56 & 68.26 & 88.18 \\
Heavy\_Makeup         & 66.06 & 85.68 & 89.69 & 92.22 & 84.18 & 87.20 & 84.43 & 91.49 & 90.01 & 91.86 \\
High\_Cheekbones      & 73.33 & 86.62 & 90.78 & 92.21 & 84.42 & 87.13 & 89.02 & 92.27 & 90.39 & 91.72 \\
Gender.               & 46.67 & 80.14 & 85.56 & 90.87 & 74.44 & 83.93 & 70.00 & 87.73 & 90.56 & 91.52 \\
Mouth\_Slightly\_Open & 74.22 & 87.01 & 91.27 & 92.33 & 84.51 & 87.42 & 91.01 & 92.56 & 91.74 & 91.85 \\
Mustache              & 50.00 & 80.89 & 91.72 & 95.38 & 50.00 & 78.58 & 91.50 & 95.97 & 91.60 & 94.93 \\
Narrow\_Eyes          & 69.23 & 85.54 & 90.05 & 91.85 & 82.94 & 87.00 & 88.46 & 91.84 & 91.69 & 91.90 \\
No\_Beard             & 39.39 & 78.10 & 84.85 & 90.97 & 72.73 & 83.80 & 57.58 & 85.00 & 84.85 & 90.43 \\
Oval\_Face            & 75.16 & 87.20 & 90.71 & 92.40 & 84.22 & 87.70 & 91.24 & 92.76 & 90.31 & 91.90 \\
Pale\_Skin            & 75.44 & 87.99 & 90.30 & 91.54 & 81.67 & 85.97 & 91.37 & 92.46 & 89.55 & 92.02 \\
Pointy\_Nose          & 73.34 & 87.18 & 91.19 & 92.42 & 84.87 & 87.69 & 89.29 & 92.55 & 91.07 & 92.00 \\
Receding\_Hairline    & 66.67 & 84.75 & 90.98 & 91.86 & 80.56 & 84.34 & 83.33 & 91.11 & 87.96 & 90.97 \\
Rosy\_Cheeks          & 74.90 & 88.17 & 91.40 & 93.32 & 84.88 & 88.59 & 90.55 & 93.00 & 91.49 & 92.71 \\
Sideburns             & 38.46 & 77.84 & 84.62 & 90.69 & 76.92 & 84.14 & 76.92 & 89.72 & 91.35 & 93.75 \\
Smiling               & 75.91 & 86.87 & 91.59 & 92.31 & 84.14 & 87.29 & 91.10 & 92.49 & 91.53 & 91.88 \\
Straight\_Hair        & 74.00 & 86.60 & 90.27 & 92.05 & 84.37 & 87.41 & 88.36 & 92.37 & 91.55 & 91.92 \\
Wavy\_Hair            & 74.22 & 86.88 & 91.41 & 92.40 & 84.15 & 87.41 & 88.81 & 92.21 & 91.64 & 91.88 \\
Wearing\_Earrings     & 75.36 & 86.77 & 91.67 & 92.63 & 84.78 & 87.70 & 90.88 & 92.55 & 91.51 & 92.19 \\
Wearing\_Hat          & 7.69  & 70.39 & 46.15 & 82.28 & 46.15 & 77.34 & 61.54 & 86.26 & 53.85 & 84.56 \\
Wearing\_Lipstick     & 59.37 & 83.61 & 89.47 & 91.88 & 82.53 & 86.01 & 79.37 & 90.10 & 90.32 & 91.57 \\
Wearing\_Necklace     & 74.57 & 87.26 & 91.09 & 92.47 & 82.26 & 87.42 & 89.53 & 92.17 & 90.73 & 92.21 \\
Wearing\_Necktie      & 25.00 & 74.52 & 90.00 & 91.56 & 80.00 & 84.31 & 35.00 & 79.32 & 91.44 & 92.53 \\
Young                 & 71.60 & 86.07 & 89.13 & 91.64 & 76.21 & 85.96 & 90.19 & 92.03 & 87.23 & 91.51 \\
\midrule \midrule
Average               & 60.06 & 83.63 & 84.25 & 90.83 & 78.51 & 85.79 & 82.52 & 90.95 & 87.04 & 91.41 \\
  \bottomrule
  \end{tabular}
  \caption{
    Worst-group and average-group accuracy for hair color prediction on CelebA.
  }\label{tab:celeba}
\end{table*}

\end{document}